%% file: _MAIN.tex
\documentclass[11pt]{article}

\usepackage[T1]{fontenc}
\usepackage[utf8]{inputenc}
\usepackage{lmodern}
\usepackage[a4paper,margin=2.5cm]{geometry}
\usepackage{amsmath,amssymb,amsfonts}
\usepackage{graphicx}
\usepackage{booktabs}
\usepackage{xcolor}
\usepackage{rotating}
\usepackage{algorithm}
\usepackage{algpseudocode}
\usepackage{microtype}
\usepackage[authoryear,round]{natbib}
\usepackage[hidelinks]{hyperref}

\newcommand{\R}{\mathbb{R}}
\newcommand{\ind}{\mathbb{I}}
\newcommand{\simplex}{\boldsymbol{\Delta}}

\newcommand{\ba}{\boldsymbol{a}}
\newcommand{\bp}{\boldsymbol{p}}
\newcommand{\be}{\boldsymbol{e}}
\newcommand{\bX}{\boldsymbol{X}}
\newcommand{\bU}{\boldsymbol{U}}
\newcommand{\bS}{\boldsymbol{S}}
\newcommand{\br}{\boldsymbol{r}}
\newcommand{\bZ}{\boldsymbol{Z}}
\newcommand{\bV}{\boldsymbol{V}}
\newcommand{\bb}{\boldsymbol{b}}
\newcommand{\bu}{\boldsymbol{u}}
\newcommand{\bG}{\boldsymbol{G}}
\newcommand{\bM}{\boldsymbol{M}}
\newcommand{\bz}{\boldsymbol{z}}
\DeclareMathOperator*{\argmin}{arg\,min}

\begin{document}

\begin{center}
{\LARGE\bfseries Global--Local and Window-Concatenated Functional Random Forests: Early and Late Fusion of Window-Specific Functional Principal Components\par}
\vspace{1.2em}
{\large Fabrizio Maturo$^{1,*,\dagger}$ \quad Annamaria Porreca$^{2,3,\dagger}$\par}
\vspace{1em}
{\small
$^{1}$Department of Economics, Statistics and Business, Faculty of Technological and Innovation Sciences, Universitas Mercatorum, Piazza Mattei 10, 00186 Rome, Italy. ORCID: 0000-0002-2362-4970\\[2pt]
$^{2}$Department for the Promotion of Human Science and Quality of Life, San Raffaele University, Via di Val Cannuta 247, 00166 Rome, Italy\\[2pt]
$^{3}$IRCCS San Raffaele Roma, Via della Pisana 235, 00163 Rome, Italy. ORCID: 0000-0003-3278-1561\\[6pt]
$^{*}$Corresponding author: Fabrizio Maturo, \href{mailto:fabrizio.maturo@unimercatorum.it}{fabrizio.maturo@unimercatorum.it}\\
Annamaria Porreca: \href{mailto:annamaria.porreca@uniroma5.it}{annamaria.porreca@uniroma5.it}\\
$^{\dagger}$These authors contributed equally to this work.\par}
\end{center}

\begin{abstract}
Functional random forests built on a single functional principal component decomposition of the whole domain may dilute features confined to short intervals, whose location is rarely known in advance. We propose two random forests based on window-specific functional principal components. The global--local functional random forest (GL-FRF) is a late-fusion method that stacks a tuned full-curve forest and one forest per window with penalised convex weights; it contains the full-curve forest as a feasible solution, and its weights indicate which parts of the domain drive the prediction. The window-concatenated functional random forest (WC-FRF) is an early-fusion method that trains a single tuned forest on the concatenated window-specific scores. On twenty UCR datasets, GL-FRF improved on a tuned full-curve forest with the same number of trees on 16, with Brier score reductions of up to 0.21, significantly after Holm adjustment. No method was best on more than six datasets. WC-FRF was more accurate than GL-FRF on 13 datasets but failed when the signal required the full curve, whereas GL-FRF had the smallest mean and maximum excess over the best method across all 32 simulated and real settings, which makes it the safer choice when the location of the signal is unknown.
\end{abstract}


\section{Introduction}\label{sec:intro}

Functional observations arise in many scientific and applied domains where measurements are recorded over time or along another ordered continuum. Examples include electrocardiographic signals, industrial sensor traces, spectrometric curves, motion trajectories and environmental monitoring profiles. In these settings, the statistical unit is naturally represented as a function rather than as a finite vector of unrelated measurements, and functional data analysis (FDA) provides an explicit representation of the domain, of the dependence between neighbouring observations, of the smoothness of the underlying trajectories and of the geometry in which curves are compared \citep{ramsay2005,ferratyvieu2006}. When the sampling grid is sufficiently informative, an individual observation can be regarded as a realisation of a random function $X\in L^2(\mathcal T)$, and supervised classification consists in predicting a categorical outcome $Y\in\{1,\ldots,C\}$ from $X$.

Supervised classification of functional data has been addressed through functional linear discriminant analysis \citep{jameshastie2001}, nonparametric kernel methods \citep{ferratyvieu2003}, supervised projections such as functional partial least squares \citep{preda2007}, and, more recently, tree-based ensembles. A key tool in most of these approaches is functional principal component analysis (FPCA), which diagonalises the covariance operator of the predictor and summarises its dominant modes of variation. FPCA, however, orders components by explained variance and not by their ability to separate the classes. Components with small eigenvalues may carry most of the discriminative information \citep{jolliffe1982,hall2001}, and projections that are suboptimal for describing the curves may be close to optimal for classifying them \citep{delaiglehall2012}. In tree-based classifiers, the problem has a second side: retaining many weakly informative components reduces the probability that the informative ones are considered at each split. For these reasons, the functional representation and the classifier should be regarded as interconnected parts of the same statistical procedure rather than as unrelated preprocessing and learning steps \citep{maturoverde2023}.

Several contributions have shown how functional structure can be combined with decision trees and ensembles. \citet{maturoverde2022} and \citet{maturoverde2023} developed functional classification trees and functional random forests (FRF) based on FPCA scores, with attention to both predictive performance and interpretability, and \citet{maturoverde2024} combined unsupervised and supervised learning to improve functional classifiers. \citet{loffredo2025} extended FPCA-based forests to censored time-to-event outcomes, while \citet{maturoadac2026} proposed explainability tools for FRF that relate predictions to individual functional principal components. \citet{gregorutti2015} studied grouped importance measures for functional variables, \citet{belli2022} introduced functional splits defined by learned integral features, and \citet{moller2016} proposed random forests whose predictors are the means of the curves over randomly generated intervals of the domain. Ensemble diversity induced at the level of the functional representation has been exploited by the functional voting classifier \citep{riccio2024}, by randomised spline trees \citep{riccio2024rst}, by augmented functional random forests that combine curves and their derivatives \citep{maturojoc2026}, and by geometric feature enrichment \citep{maturoasta2026}. These contributions show that the functional representation itself, and not only the choice of classifier, influences what an ensemble can learn.

A complementary literature focuses on locality in the functional domain. \citet{liyu2008} proposed a segmentation approach to functional classification, in which discriminant features are extracted from segments of the domain; \citet{berrendero2016} addressed the selection of the most informative points of the trajectory; and \citet{pinivantini2017} developed interval-wise inference to identify the parts of the domain on which functional populations differ. On the representation side, localised FPCA \citep{chenlei2015} estimates eigenfunctions supported on subintervals, and multiscale FPCA \citep{linzhu2019} partitions the domain according to the scale of the variance and performs FPCA on each subdomain, precisely because features located in regions of low variance can be concealed by a single decomposition of the whole domain. In time series classification, interval-based forests summarise random intervals through simple statistics \citep{deng2013,baydogan2013}; the Canonical Interval Forest and its diverse-representation variant are among the most accurate interval methods \citep{middlehurst2020cif,middlehurst2021hc2}, and the bake-off of \citet{bagnall2017} has shaped the evaluation practice in this field.

From an FDA perspective, an interval is not simply a subset of columns: it is a restriction of the functional argument, with its own mean, covariance operator, smoothing requirements, eigenfunctions and explained-variance spectrum. An FPCA carried out separately on an interval can therefore expose modes of variation that would not appear among the leading components of a global FPCA. At the same time, independent local models cannot in general reconstruct predictive interactions between distant intervals: if class membership depends on the joint configuration of two separated features, each feature may be marginally uninformative while their interaction is perfectly predictive. These two observations raise a simple question, which motivates the present work: should prediction be entrusted to a single functional random forest trained on the whole trajectory, or to a combination of forests whose functional representations are restricted to different intervals, and under which conditions does the second option improve on the first?

Window-specific FPCA has been used before. \citet{yan2018} applied FPCA to moving time windows of longitudinal biomarkers to update predictions of times to event, and localised FPCA \citep{chenlei2015} estimates eigenfunctions with local support rather than decomposing windows fixed in advance. The closest antecedent is multiscale FPCA \citep{linzhu2019}, which divides the domain into subdomains according to the level of the variance function, performs a separate FPCA on each of them and, in an application to diffusion tensor imaging profiles, uses the resulting scores as inputs of a random forest to classify patients with multiple sclerosis and healthy controls, with a small improvement over single-scale FPCA. Subdomain FPCA scores have therefore already been used as features of a random forest, and the untuned forest on concatenated window-specific scores considered in this paper, WC-0, can be regarded as a version of that approach with equal-width windows. To the best of our knowledge, what has not been studied is how window-specific components should be combined with the full-curve representation in a functional random forest, whether early and late fusion of the same window-specific representation behave differently, and whether the gains over a tuned full-curve forest are due to localisation or to the diversity of the resulting ensemble.

This paper proposes and compares two functional random forests built on the same window-specific representation, which differ in how the information of the windows is combined. In the terminology of multi-view learning, the first is a late-fusion and the second an early-fusion strategy. The \emph{global--local functional random forest} (GL-FRF) retains a full-curve FRF, tuned by internal cross-validation, alongside $W$ local FRFs, each trained on the FPCA scores of a separately smoothed and separately decomposed window. Their predicted class probabilities are combined through convex weights estimated from cross-fitted predictions, with two penalties that shrink the total local weight towards the global expert and the local weights towards equality, and the number of windows and the penalties are selected by nested cross-validation. Because the weights lie on the probability simplex, the global expert is a feasible solution and GL-FRF reduces exactly to a tuned full-curve forest when the local experts do not improve the cross-fitted loss; the estimated weights also indicate which parts of the domain contribute to the prediction. The \emph{window-concatenated functional random forest} (WC-FRF) concatenates the window-specific scores into a single feature vector and trains one forest whose number of candidate variables at each split and minimum node size are tuned on the out-of-bag Brier score. WC-FRF lets the forest select, split by split, the windows that matter and can model interactions between windows, but it has no access to the full-curve representation and no explicit fallback.

The contributions of the article are the following. First, building on multiscale FPCA \citep{linzhu2019}, we formulate window-specific functional principal components as a representation for functional random forests, with smoothing, FPCA and all tuning carried out within the training data, and we propose GL-FRF and WC-FRF as two ways of exploiting it. GL-FRF is a structured application of stacking \citep{wolpert1992,breiman1996,vanderlaan2007} to functional random forests \citep{breiman2001}, with weights fixed after training, unlike query-specific dynamic ensemble selection \citep{cruz2015}; WC-FRF differs from the random forest on multiscale FPCA scores of \citet{linzhu2019} in the equal-width windows, the separate smoothing of each window and the out-of-bag tuning of the forest, and it replaces the interval features of the forests of \citet{moller2016} and \citet{deng2013}, namely means, standard deviations and slopes, with functional principal components of each window. Second, we propose a locality control to separate the contribution of temporal localisation from that of ensemble diversity, which are usually confounded in ensembles of functional representations \citep{riccio2024}: the same stacking procedure is applied to random non-contiguous partitions of the grid with the same set sizes as the windows. Third, we characterise when each method is preferable, by comparing both with tuned full-curve forests, random-interval forests and functional nearest neighbours under a common protocol. Fourth, the methods are evaluated on twenty benchmark datasets with repeated stratified resampling, and the main comparisons are tested across datasets with an adjustment for multiplicity.

The results show that the two methods are complementary and that neither should be judged only by the number of datasets on which it is the most accurate. No method among the twelve compared was the best on more than six of the twenty datasets, and the most accurate method changed with the dataset in a way that is difficult to anticipate without knowing where the discriminative information lies; in this situation, the relevant property of a method is how far it can fall from the best. On real curves, window-specific components carry information that a tuned full-curve decomposition does not represent, and the locality control indicates that this information is mainly due to localisation rather than to the diversity of the ensemble. WC-FRF exploits it more accurately than GL-FRF on most real datasets, but fails when the signal requires the full curve or an interaction captured by the global representation. GL-FRF is designed as a safe default rather than as the most accurate method on every dataset: it behaves like a tuned full-curve forest when the windows are uninformative, exploits the windows when they are informative, and had the smallest mean and maximum distance from the best method across all simulated and real settings. Its weights also indicate, for each dataset, whether and where the windows matter.

The paper is organised as follows. Section~\ref{sec:prelim} recalls the functional representation and FPCA-based random forests. Section~\ref{sec:method} introduces GL-FRF and WC-FRF, their algorithms and their properties. Section~\ref{sec:design} describes the experimental design, including the locality control and the competitors. Section~\ref{sec:sim} reports the simulation study and Section~\ref{sec:real} the analysis of the benchmark datasets. Section~\ref{sec:discussion} discusses the results, their limitations and possible extensions.

\section{Preliminaries}\label{sec:prelim}

\subsection{Functional data and penalised reconstruction}\label{sec:framework}

Let $(X_i,Y_i)$, $i=1,\ldots,n$, be independent pairs, where $X_i\in L^2(\mathcal T)$ is a scalar functional predictor on a compact interval $\mathcal T=[a,b]$ and $Y_i\in\mathcal C=\{1,\ldots,C\}$, $C\geq2$. In practice, each curve is observed on a common equally spaced grid $a\leq t_1<\cdots<t_p\leq b$, and the first step of the analysis consists in reconstructing a smooth function from the discrete observations \citep{ramsay2005}. Throughout the paper, every quantity estimated from data, namely smoothing parameters, means, principal components, forests, aggregation weights and tuning parameters, is computed on a training set only; held-out curves are projected onto training-based representations and used for evaluation. For a training set $D$, $n_D$ denotes its size.

For a domain $I$, which is either the full grid or a window of it, the observations of curve $i$ restricted to $I$, $\boldsymbol x_{i,I}$, are smoothed by cubic B-splines $\{B_{I\ell}\}_{\ell=1}^{q_I}$ defined on that domain:
\begin{equation}\label{eq:smoothing}
 \widehat X_{i,I}(t)=\sum_{\ell=1}^{q_I}\widehat c_{i,I\ell}B_{I\ell}(t),\qquad
 \widehat{\boldsymbol c}_{i,I}=\argmin_{\boldsymbol c}
 \left\{\|\boldsymbol x_{i,I}-\mathbf B_I\boldsymbol c\|^2
 +\lambda_I\|\mathbf D_2\boldsymbol c\|^2\right\}.
\end{equation}
Following the P-spline approach \citep{eilersmarx1996}, the basis is large, $q_I=\min\{100,|I|-1\}$ for $|I|\geq5$, and smoothness is controlled by the penalty, where $\mathbf B_I$ is the $|I|\times q_I$ matrix of basis functions evaluated at the grid points of $I$, with equally spaced knots, and $\mathbf D_2$ is the $(q_I-2)\times q_I$ second-order difference matrix of the coefficients. The smoothing parameter belongs to $\{10^{-6},10^{-5},\ldots,10^{4}\}$ and is selected on the current training set by pooled generalised cross-validation, using predictor measurements only. The resulting linear smoother is applied without modification to held-out curves, and a window with fewer than five grid points is left unsmoothed. Smoothing is repeated within all cross-fitting, model-selection and validation folds.

\subsection{Functional principal components}\label{sec:fpca}

For a training set $D$ and an index set $I\subseteq\{1,\ldots,p\}$, let $\bX_{D,I}$ be the $n_D\times|I|$ matrix of reconstructed functional values and $\widehat\mu_{D,I}$ its vector of column means. On an equally spaced grid, FPCA reduces, up to a constant quadrature factor that does not affect the explained-variance ratios, to the singular value decomposition
\begin{equation}\label{eq:svd}
 \bZ_{D,I}=\bX_{D,I}-\boldsymbol{1}\widehat\mu_{D,I}^{\top}
 =\bU_{D,I}\bS_{D,I}\bV_{D,I}^{\top}.
\end{equation}
With squared singular values $s_1^2\geq\cdots\geq s_r^2>0$ and an explained-variance threshold $\tau\in(0,1]$, the number of retained components is
\begin{equation}\label{eq:pcak}
 K_{D,I}=\min\left\{k\geq 1:\ \frac{\sum_{h=1}^{k}s_h^2}{\sum_{h=1}^{r}s_h^2}\geq \tau\right\},
\end{equation}
and the vector of scores of a curve $x$ restricted to $I$ is
\begin{equation}\label{eq:score}
 \bz_{D,I}(x)=\bV_{D,I,K}^{\top}(\widehat{\boldsymbol x}_I-\widehat\mu_{D,I})\in\R^{K_{D,I}},
\end{equation}
where $\widehat{\boldsymbol x}_I$ is the vector of values on $I$ of the curve reconstructed with the smoother estimated on $D$, and $\bV_{D,I,K}$ contains the first $K_{D,I}$ right singular vectors. The thresholds used for the global and the window-specific decompositions are given in Section~\ref{sec:experts}. Test curves are centred with the training mean and projected onto the training loadings. Curves are not subject to additional amplitude standardisation.

\subsection{FPCA-based functional random forests}\label{sec:frf}

An FPCA-based functional random forest \citep{maturoverde2022,maturoverde2023} is a random forest \citep{breiman2001} trained on the score vectors $\bz_{D,I}(X_i)$. In this work all forests are probability forests implemented in \texttt{ranger} \citep{wright2017}, grown with bootstrap sampling and the default number $\lfloor\sqrt K\rfloor$ of candidate variables at each split, where $K$ is the number of input scores. Each forest returns a vector of estimated class probabilities
\begin{equation}\label{eq:experts}
 \widehat\bp(x)=\left(\widehat P(Y=1\mid x),\ldots,\widehat P(Y=C\mid x)\right)^\top\in\simplex_C,
 \qquad \simplex_C=\Bigl\{\bp\in[0,1]^C:\textstyle\sum_{c}p_c=1\Bigr\}.
\end{equation}

The choice of $\tau$ plays a central role. When FPCA is used to describe the main modes of variation, thresholds of 70--80\% are common, but in classification the components that separate the classes often explain a small share of the variability, because the leading components capture variation shared by all classes \citep{jolliffe1982,maturoverde2023,maturoverde2024}. Conversely, a high threshold on a long domain retains many weakly informative scores, which a forest with $\lfloor\sqrt K\rfloor$ candidate variables at each split rarely considers, and which can degrade the sharpness of the predicted probabilities. In GL-FRF the two risks are handled differently on the full domain and on the windows, as described in Section~\ref{sec:method}.

\section{Functional random forests on window-specific components}\label{sec:method}

\subsection{Motivation}\label{sec:motivation}

For $W\in\mathcal W=\{5,10,20\}$, the grid is partitioned into $W$ contiguous windows with index sets
\begin{equation}\label{eq:windows}
 I_j(W)=\left\{\left\lfloor\frac{(j-1)p}{W}\right\rfloor+1,\ldots,
 \left\lfloor\frac{jp}{W}\right\rfloor\right\},\qquad j=1,\ldots,W,
\end{equation}
whose sizes differ by at most one when $p\geq W$. Let $\widehat\bp_0$ be the probability vector of a forest trained on the global scores and $\widehat\bp_j$, $j=1,\ldots,W$, that of a forest trained only on the scores of window $I_j(W)$.

Two simple situations illustrate why neither the global nor the local representation should be preferred a priori. If the restriction $X|_{I_j}$ is independent of $Y$ for every $j$, the posterior class probabilities based on any single window equal the class prior, although the full trajectory may predict $Y$ exactly through an interaction between windows. For instance, if $Y=\ind(AB=1)$ with $A$ and $B$ independent Rademacher variables encoded in two different windows,
\begin{equation}\label{eq:xor_structure}
 P(Y=1\mid A)=P(Y=1\mid B)=\tfrac12,
 \qquad P(Y=1\mid A,B)=\ind(AB=1),
\end{equation}
and every convex combination of window-specific posteriors equals the prior, whereas a classifier with access to both windows can recover $Y$. Conversely, when the discriminating signal is a short transient, or is located in a part of the domain where the curves vary little, a global FPCA may spread it over many components of small variance \citep{linzhu2019}, which a forest considers infrequently; a window-specific FPCA can expose it among its leading components. These two situations motivate a combination in which the global forest is always retained and the local forests receive weight only when the training data support it.

\subsection{Experts}\label{sec:experts}

For a training set $D$ and a number of windows $W$, the global expert $\mathcal F_{0,D}$ and the local experts $\mathcal F_{j,D,W}$, $j=1,\ldots,W$, are probability forests with $B=100$ trees each, trained respectively on the global and on the window-specific FPCA scores. Each window is smoothed and decomposed separately, so that no information from other parts of the domain enters a nominally local expert, and local scores are never obtained by slicing global components.

On each window, which is a short subinterval with a rapidly decaying spectrum, the local experts retain the components that reach 99\% of the explained variance, $\tau_{\mathrm{loc}}=0.99$, and use minimum node size one. On the full domain, the threshold $\tau_0$ and the minimum node size $m_0$ of the global expert are not fixed: they are selected from $\{0.80,0.90,0.95,0.99\}\times\{1,5,10\}$ by three-fold stratified cross-validation of the multiclass Brier score on $D$. Within each fold the smoother and the FPCA are refitted on the fold's training curves, and all candidate thresholds use the same decomposition. Exact ties are broken in favour of the larger node size and then of the lower threshold, and the selected configuration is refitted on $D$.

\subsection{Cross-fitted level-one predictions}\label{sec:crossfit}

The training set is divided into three class-stratified folds $D_1,D_2,D_3$. For each $v$, all representations and all $W+1$ forests are refitted on $D\setminus D_v$, including the tuning of the global expert, and used to predict the curves in $D_v$, giving
\begin{equation}\label{eq:oof}
 \widetilde\bp_{ij,W}=\widehat\bp_{j,D\setminus D_{v(i)},W}(X_i),
 \qquad i\in D,\ j=0,\ldots,W,
\end{equation}
where $v(i)$ is the fold containing $i$. All experts share the fold assignment, so that no prediction $\widetilde\bp_{ij,W}$ uses observation $i$. These out-of-fold predictions are used only to estimate the aggregation weights; the experts used for prediction are then refitted on the whole of $D$.

\subsection{Penalised convex aggregation}\label{sec:aggregation}

Let $\ba_W=(a_0,\ldots,a_W)^\top\in\simplex_{W+1}$ and let $\eta(\ba_W)=1-a_0=\sum_{j=1}^Wa_j$ denote the total local weight. The aggregated prediction is
\begin{equation}\label{eq:mixture}
 \widehat\bp_{\mathrm{GL},W}(x;\ba_W)=\sum_{j=0}^{W}a_j\widehat\bp_j(x)
 =(1-\eta)\,\widehat\bp_0(x)+\eta\sum_{j=1}^W w_j\widehat\bp_j(x),
 \qquad w_j=a_j/\eta\ \ (\eta>0).
\end{equation}
With $\be_i$ the one-hot vector of $Y_i$, the cross-fitted multiclass Brier loss is
\begin{equation}\label{eq:brier}
 \widehat L_D(\ba_W)=\frac1{n_D}\sum_{i\in D}
 \Bigl\|\be_i-\sum_{j=0}^{W}a_j\widetilde\bp_{ij,W}\Bigr\|_2^2,
\end{equation}
and the penalty is
\begin{equation}\label{eq:penalty}
 \Omega_W(\ba_W;\lambda_G,\lambda_L)=
 \lambda_G\,\eta(\ba_W)^2+
 \lambda_L\sum_{j=1}^{W}\Bigl(a_j-\frac{\eta(\ba_W)}{W}\Bigr)^2,
 \qquad\lambda_G,\lambda_L\geq 0.
\end{equation}
The first term shrinks the aggregate towards the global forest, while the second shrinks the local weights towards equality without fixing them. The weights are
\begin{equation}\label{eq:optim}
 \widehat\ba_W(\lambda_G,\lambda_L)=
 \argmin_{\ba_W\in\simplex_{W+1}}
 \bigl\{\widehat L_D(\ba_W)+\Omega_W(\ba_W;\lambda_G,\lambda_L)\bigr\},
\end{equation}
which is a convex quadratic programme with linear constraints, solved by projected gradient descent (Section~\ref{sec:properties}). A new curve is assigned to $\arg\max_c[\widehat\bp_{\mathrm{GL},W}(x;\widehat\ba_W)]_c$, with ties broken by the first class in a fixed ordering. Because the weights are constant across test curves, GL-FRF is a global stacking rule and not a dynamic, query-specific selector \citep{cruz2015}. The unweighted Brier loss is preferred to a class-balanced version so that the fitted probabilities refer to the class distribution of the training population; balanced accuracy is reported alongside it because several datasets are imbalanced.

When $\lambda_G=\lambda_L=0$, problem~\eqref{eq:optim} reduces to the unpenalised simplex stack of the same experts, which we denote Stack-0. Since $\lambda_G=\lambda_L=0$ belongs to the grid of Section~\ref{sec:selection}, GL-FRF includes Stack-0 as a special case, and the comparison between the two quantifies what the penalties add to cross-fitted convex stacking.

\subsection{Window-concatenated functional random forest}\label{sec:wcfrf}

WC-FRF uses the same window-specific representation as the local experts of GL-FRF, but combines it before the forest rather than after it. For a training set $D$ and $W$ windows, the score vectors of all windows are concatenated,
\begin{equation}\label{eq:wc}
 \bz^{\mathrm{WC}}_{D,W}(x)=\bigl(\bz_{D,I_1(W)}(x)^\top,\ldots,\bz_{D,I_W(W)}(x)^\top\bigr)^\top\in\R^{K_W},
 \qquad K_W=\sum_{j=1}^W K_{D,I_j(W)},
\end{equation}
where each window is smoothed by \eqref{eq:smoothing} with its own smoothing parameter and decomposed with $\tau_{\mathrm{loc}}=0.99$, exactly as in Section~\ref{sec:experts}. A single probability forest with $100(W+1)$ trees, the total number of trees of GL-FRF, is trained on $\bz^{\mathrm{WC}}_{D,W}$. The number of candidate variables at each split is chosen from $\{\lfloor\sqrt{K_W}\rfloor,\lfloor K_W/5\rfloor,\lfloor K_W/3\rfloor,\lfloor K_W/2\rfloor\}$ and the minimum node size from $\{1,5\}$ by the out-of-bag Brier score on $D$, because the default value $\lfloor\sqrt{K_W}\rfloor$ is small when many windows are concatenated and the number of candidate variables is among the hyperparameters that most affect random forests \citep{probst2019}. In this study WC-FRF uses the number of windows $\widehat W$ selected for GL-FRF by the nested procedure of Section~\ref{sec:selection}.

The two methods have complementary properties. WC-FRF can split on the scores of different windows within the same tree, and can therefore represent interactions between windows that a convex combination of window-specific posteriors cannot, but it does not include the full-curve representation, so that information spread over the domain must be reconstructed from local components. GL-FRF retains the full-curve forest and can reduce to it exactly (Property 1 below), and its weights summarise the contribution of each window, but each local expert sees one window only and interactions between windows are available only through the global expert.

\subsection{Selection of $W$, $\lambda_G$ and $\lambda_L$}\label{sec:selection}

The grids are $\mathcal W=\{5,10,20\}$ and $\Lambda_G=\Lambda_L=\{0,10^{-3},10^{-2},10^{-1},1\}$. Selection uses nested cross-validation within the training data. The training set is split into three stratified inner folds; for each configuration and inner fold, the whole method (representations, tuning of the global expert, experts, a separate three-fold cross-fitting layer, and weights) is fitted on the inner training part, and its Brier loss is computed on the inner validation part. The configuration with the smallest mean inner loss is selected, with exact ties broken in favour of fewer windows, then larger $\lambda_G$, then larger $\lambda_L$. The method is then refitted on the full training set. The nested design requires every class to contain at least $3\times(3+1)=12$ training curves, so that each inner training part can itself be cross-fitted without losing a class; datasets that do not meet this condition are excluded and reported.

\subsection{Algorithm}\label{sec:algorithm}

Algorithm~\ref{alg:glfrf} summarises the complete procedure. The function \texttt{FitExperts} is called at every level of the nested scheme, so that smoothing, FPCA, tuning of the global expert and forests are always estimated on the curves available at that level.

\begin{algorithm}[t]
\caption{GL-FRF: fitting on a training set and prediction of test curves}\label{alg:glfrf}
\begin{algorithmic}[1]
\Function{FitExperts}{$D$, $W$}
  \State Select $(\widehat\tau_0,\widehat m_0)$ by three-fold stratified CV of the Brier score on $D$, refitting the smoother and the FPCA on the training part of each fold
  \State Global domain: select $\lambda$ by GCV on $D$, smooth the curves of $D$ by \eqref{eq:smoothing}, fit FPCA with $\widehat\tau_0$ and a forest with $B$ trees and node size $\widehat m_0$
  \For{$j=1,\ldots,W$}
    \State Window $I_j(W)$: smooth by \eqref{eq:smoothing} with its own GCV; fit FPCA with $\tau_{\mathrm{loc}}=0.99$ and a forest with $B$ trees and node size one
  \EndFor
  \State \Return smoothers, FPCA means and loadings, forests
\EndFunction
\Function{CrossFit}{$D$, $W$}
  \State Split $D$ into three stratified folds $D_1,D_2,D_3$
  \For{$v=1,2,3$}
    \State $\mathcal E\gets$ \Call{FitExperts}{$D\setminus D_v$, $W$}; store the predictions $\widetilde\bp_{ij,W}$ of $\mathcal E$ for $i\in D_v$
  \EndFor
  \State \Return $\{\widetilde\bp_{ij,W}\}$
\EndFunction
\Require training set $D_{\mathrm{tr}}$ with at least 12 curves per class, test curves $D_{\mathrm{te}}$, $\mathcal W$, $\Lambda_G$, $\Lambda_L$, $B=100$
\State Split $D_{\mathrm{tr}}$ into three stratified inner folds $V_1,V_2,V_3$
\For{$W\in\mathcal W$ and $v=1,2,3$}
  \State $\widetilde P\gets$ \Call{CrossFit}{$D_{\mathrm{tr}}\setminus V_v$, $W$}; $\ \mathcal E\gets$ \Call{FitExperts}{$D_{\mathrm{tr}}\setminus V_v$, $W$}
  \For{$(\lambda_G,\lambda_L)\in\Lambda_G\times\Lambda_L$}
    \State Solve \eqref{eq:optim} with $\widetilde P$; record the Brier loss of \eqref{eq:mixture} on $V_v$ using $\mathcal E$
  \EndFor
\EndFor
\State Select $(\widehat W,\widehat\lambda_G,\widehat\lambda_L)$ by the mean inner loss
\State $\widetilde P\gets$ \Call{CrossFit}{$D_{\mathrm{tr}}$, $\widehat W$}; solve \eqref{eq:optim} to obtain $\widehat\ba$
\State $\mathcal E\gets$ \Call{FitExperts}{$D_{\mathrm{tr}}$, $\widehat W$}
\State For each $x\in D_{\mathrm{te}}$: apply the training smoothers and loadings, compute the expert probabilities with $\mathcal E$ and return $\widehat\bp_{\mathrm{GL},\widehat W}(x;\widehat\ba)$
\end{algorithmic}
\end{algorithm}

\subsection{Properties}\label{sec:properties}

The following properties follow directly from the construction and clarify what the procedure guarantees and what it does not.

\paragraph{Property 1 (admissibility and exact fallback).}
For every $\ba_W\in\simplex_{W+1}$, $\widehat\bp_{\mathrm{GL},W}(x;\ba_W)\in\simplex_C$, and the vertex $\ba_W=(1,0,\ldots,0)^\top$ reproduces the tuned global expert exactly.

\paragraph{Property 2 (cross-fitted loss).}
For fixed $W$ and $\lambda_G,\lambda_L\geq0$, since $\Omega_W\geq0$ and $\Omega_W$ vanishes at the global vertex,
\begin{equation}\label{eq:empirical}
 \widehat L_D(\widehat\ba_W)\leq\widehat L_D(\widehat\ba_W)+\Omega_W(\widehat\ba_W;\lambda_G,\lambda_L)
 \leq \widehat L_D((1,0,\ldots,0)^\top).
\end{equation}
The inequality concerns the out-of-fold predictions used to estimate the weights. It does not imply a lower Brier score, or a lower error rate, on new observations.

\paragraph{Property 3 (sufficient condition for the global solution).}
Let $\br_i=\be_i-\widetilde\bp_{i0,W}$ and $\bu_{ij}=\widetilde\bp_{ij,W}-\widetilde\bp_{i0,W}$. If
\begin{equation}\label{eq:fallback}
 \sum_{i\in D}\br_i^{\top}\bu_{ij}\leq0\qquad
 \text{for all }j=1,\ldots,W,
\end{equation}
the global vertex minimises \eqref{eq:optim} for all $\lambda_G,\lambda_L\geq0$. Indeed, at the vertex the directional derivative of $\widehat L_D$ along the edge towards expert $j$ equals $-2n_D^{-1}\sum_i\br_i^\top\bu_{ij}\geq0$, both penalty terms have zero derivative, every feasible direction is a nonnegative combination of edge directions, and convexity gives global optimality. The condition is sufficient but not necessary, and sampling variability can produce positive local weights even when the local experts carry no information in the population.

\paragraph{Property 4 (uniqueness).}
Writing $a_j=\eta/W+d_j$ with $\sum_jd_j=0$, the penalty becomes $\lambda_G\eta^2+\lambda_L\sum_jd_j^2$, which is positive definite in $(\eta,\boldsymbol d)$, on the subspace $\sum_jd_j=0$, when $\lambda_G,\lambda_L>0$. The objective is then strictly convex on the simplex and its minimiser is unique. With $\lambda_G=\lambda_L=0$ the minimiser need not be unique when expert predictions are collinear.

\paragraph{Optimisation.}
Eliminating $a_0=1-\sum_{j\geq1}a_j$, problem \eqref{eq:optim} becomes the minimisation over $\ba_L=(a_1,\ldots,a_W)^\top$, with $\ba_L\geq\boldsymbol0$ and $\boldsymbol1^\top\ba_L\leq1$, of
\begin{equation}\label{eq:quadratic_matrix}
 f(\ba_L)=\ba_L^{\top}\bG\ba_L-2\bb^{\top}\ba_L,
 \quad\bG=\frac{\bM^{\top}\bM}{n_D}+\lambda_G\boldsymbol{1}\boldsymbol{1}^{\top}+
 \lambda_L\Bigl(\boldsymbol{I}-\frac{\boldsymbol{1}\boldsymbol{1}^{\top}}{W}\Bigr),
 \quad\bb=\frac{\bM^{\top}\boldsymbol{\rho}}{n_D},
\end{equation}
up to a constant, where the $j$th column of $\bM$ stacks the vectors $\bu_{ij}$ of Property 3 over $i\in D$ and $\boldsymbol\rho$ stacks the residuals $\br_i$. The penalty terms follow from $\eta^2=\ba_L^\top\boldsymbol1\boldsymbol1^\top\ba_L$ and $\sum_j(a_j-\eta/W)^2=\ba_L^\top(\boldsymbol I-\boldsymbol1\boldsymbol1^\top/W)\ba_L$. The implementation uses projected gradient descent with step $1/L$, $L=2\lambda_{\max}(\bG)$, and exact Euclidean projection onto $\{\ba_L\geq\boldsymbol0,\ \boldsymbol1^\top\ba_L\leq1\}$. Iterations stop when the largest coordinate change is below $10^{-9}$ or after 12,000 iterations, and convergence is recorded for every fit. The projection allows exact zero weights and therefore the exact global solution.

\paragraph{Computational cost.}
Each call of \texttt{FitExperts} fits $W+1$ forests and, for the global expert, 36 additional 100-tree forests for the $4\times3$ tuning grid and three folds; the final fit calls it four times, and the nested selection multiplies this by the number of candidate window counts and inner folds. Tuning results are reused when the same training set recurs across window counts. Computing times are reported in Section~\ref{sec:real}.

\section{Experimental design}\label{sec:design}

\subsection{Locality control}\label{sec:control}

A stack of $W+1$ forests trained on different representations may outperform a single forest for two different reasons. The windows may expose localised information that the global decomposition does not represent well, or the additional forests may simply make partially independent errors, so that averaging reduces variance, as in ensembles of diverse functional representations \citep{riccio2024,riccio2024rst}. Only the first explanation supports the use of temporal windows.

To separate the two explanations, we define a locality control, denoted Scr-stack. For a given $W$, the grid indices are randomly permuted and split into $W$ non-contiguous sets with the same sizes as the windows in \eqref{eq:windows}. Each set is extracted from the globally smoothed training curves, decomposed by FPCA with $\tau_{\mathrm{loc}}=0.99$ and used to train a forest with 100 trees. The global expert, the cross-fitting folds and the penalised aggregation are exactly those of GL-FRF, and the values $(\widehat W,\widehat\lambda_G,\widehat\lambda_L)$ selected for GL-FRF are used without a further nested selection. Scr-stack therefore has the same number of experts, the same number of grid points per expert and the same hyperparameters as GL-FRF, but its local experts carry no temporal locality. Using the hyperparameters of GL-FRF makes the comparison sharper and reduces the computational cost by an order of magnitude. The two stacks differ in the partition of the grid and also in the reconstruction of the curves, because a non-contiguous set of grid points cannot be smoothed as a function and Scr-stack therefore uses the globally smoothed curves, whereas each window of GL-FRF is smoothed separately. If GL-FRF and Scr-stack perform similarly, the gains of GL-FRF are attributable to ensemble diversity rather than to locality; if GL-FRF is clearly better, the result supports locality, although part of the difference may be due to the window-specific smoothing.

\subsection{Competing methods}\label{sec:competitors}

The two proposed methods, GL-FRF and WC-FRF, are compared with each other and with the following methods, all fitted on the same training data. Methods that depend on windows use the number of windows $\widehat W$ selected for GL-FRF.
\begin{itemize}
\item \emph{Ablations of the proposed methods.} Stack-0 is GL-FRF without penalties ($\lambda_G=\lambda_L=0$); WC-0 is WC-FRF with the default number of candidate variables and minimum node size one; Scr-stack is the locality control of Section~\ref{sec:control}; L-eq is the equal-weight soft vote of the local experts.
\item \emph{Global forests.} These are FPCA-based functional random forests on the full curve, as in \citet{maturoverde2022,maturoverde2023}, and represent the standard against which the window-specific representation is assessed. G-tuned is the tuned global expert of GL-FRF used alone (100 trees); G-tuned-bud is the same configuration refitted with $100(\widehat W+1)$ trees, the total number of trees of GL-FRF and WC-FRF; G-99 is an untuned global forest with $\tau_0=0.99$, minimum node size one and 100 trees, which shows the effect of tuning the global representation.
\item \emph{Random-interval forests.} IF-mean and IF-TSF consist of 20 groups of 25 trees; each group uses $\lfloor\sqrt p\rfloor$ random intervals of length at least three. IF-mean uses the interval means, as in the functional random forest of \citet{moller2016}, and IF-TSF uses the mean, standard deviation and slope of each interval, as in the time series forest \citep{deng2013}.
\item \emph{Functional nearest neighbours.} FkNN classifies a curve by the class proportions among its $k$ nearest training curves in the $L^2$ distance between globally smoothed curves, with $k\in\{1,3,5,7,9,11,15,21\}$ selected by the leave-one-out Brier score.
\end{itemize}
The comparison with G-tuned and G-tuned-bud measures what the window-specific representation adds to a tuned full-curve forest; the comparison of GL-FRF with Scr-stack measures what temporal locality adds to representation diversity; the comparison between GL-FRF and WC-FRF contrasts late and early fusion of the same representation; the interval forests and FkNN place both methods among established alternatives. The deep and dictionary-based classifiers of the time series literature \citep{middlehurst2021hc2} are not included, because the aim of the study is to assess functional representation strategies within FPCA-based forests rather than to compete with state-of-the-art time series classifiers.

\subsection{Evaluation criteria}\label{sec:criteria}

The primary criterion is the held-out multiclass Brier score, $n_{\mathrm{te}}^{-1}\sum_i\|\be_i-\widehat\bp(X_i)\|_2^2$, a strictly proper scoring rule \citep{gneiting2007} ranging from 0 to 2; an uninformative prediction $(0.5,0.5)$ gives 0.5 for two balanced classes. Secondary criteria are balanced accuracy, accuracy, macro-F1 and log loss. For every scenario or dataset and every method we compute the paired difference with GL-FRF within each replicate or split (method minus GL-FRF), so that positive differences favour GL-FRF, and the paired difference between WC-FRF and GL-FRF is reported separately. To summarise performance across heterogeneous settings, we also compute, for each scenario or dataset, the excess of each method's mean Brier score over the best of the twelve methods. The selected number of windows, the weight $\widehat a_0$ of the global expert, the selected threshold of the global expert, the convergence of the weight optimisation and the computing time are also recorded.

\section{Simulation study}\label{sec:sim}

\subsection{Simulation design}\label{sec:simdesign}

\begin{figure}[!p]
\centering
\includegraphics[width=\linewidth]{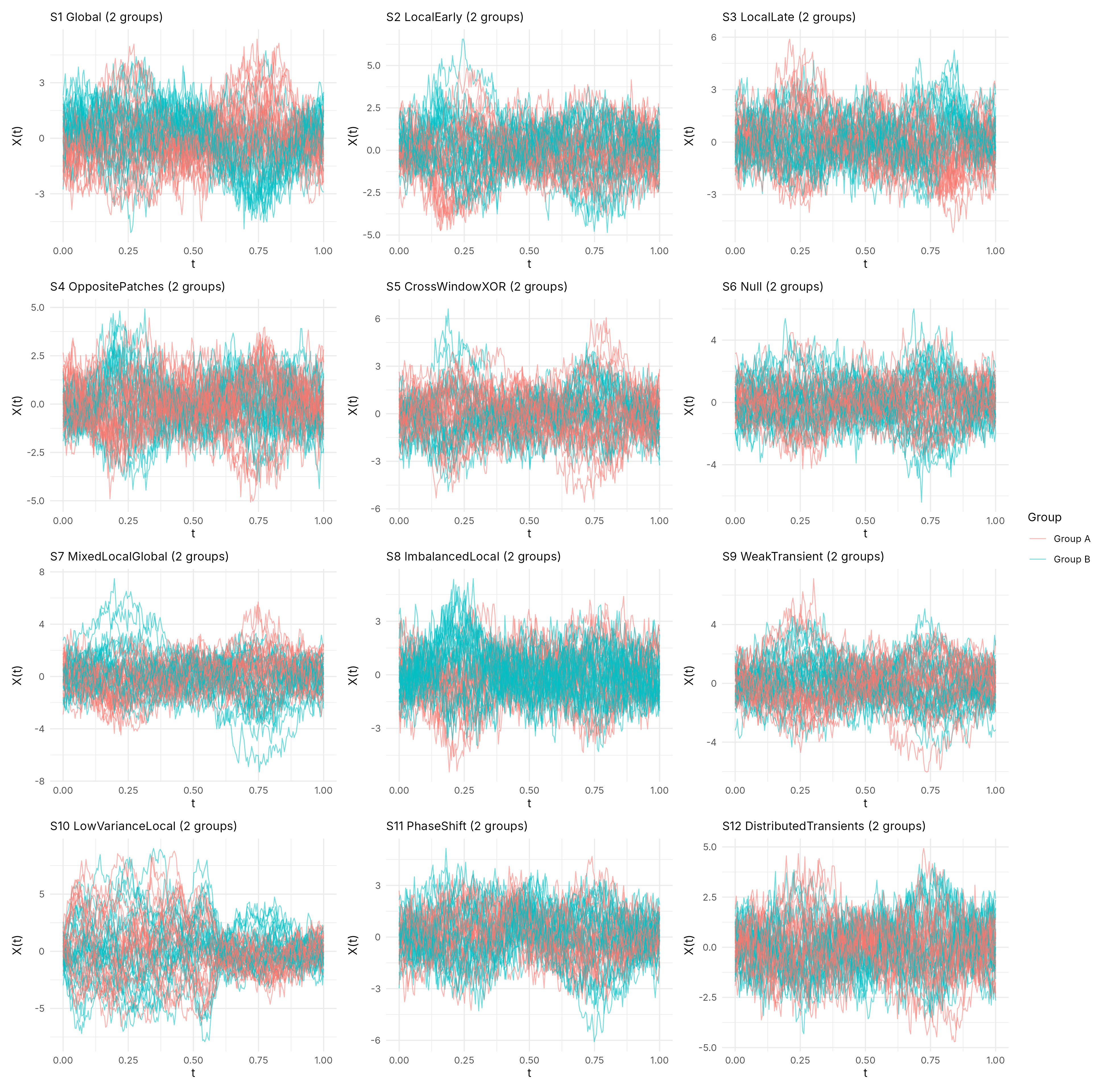}
\caption{Simulated training datasets for the twelve scenarios (first replicate). Each panel displays 25 training curves per group, with colours indicating the outcome group.}
\label{fig:simcurves}
\end{figure}

The simulation study is designed to assess GL-FRF under controlled conditions in which the location, extent, interaction structure and phase of the discriminative signal are known. Each replicate generates $n=440$ curves on $p=240$ equally spaced points in $[0,1]$, the first 240 forming the training set and the remaining 200 the test set. In all scenarios the curves share the nuisance structure
\begin{equation}\label{eq:dgp}
 x_i(t)=u_{1i}\sin(2\pi t)+u_{2i}\cos(4\pi t)+0.7\,\varepsilon_i(t)+s_i(t),
\end{equation}
where $\varepsilon_i$ is a stationary Gaussian AR(1) process on the grid with unit variance and lag-one autocorrelation $0.6$, and $s_i(t)$ is the class signal. Unless otherwise stated, $u_{1i}\sim N(0,1.4^2)$ and $u_{2i}\sim N(0,0.9^2)$, the two classes are balanced (in Scenario 5, in expectation), $z_i=\pm1$ denotes the class, and $\phi(t;\mu,\sigma)=\exp\{-(t-\mu)^2/(2\sigma^2)\}$. Each scenario is replicated $R=20$ times with replicate-specific seeds, and all methods receive identical data within a replicate. Table~\ref{tab:dgp} summarises the class signals and Figure~\ref{fig:simcurves} shows training curves from the first replicate of each scenario.

\begin{table}[t]
\centering
\small
\resizebox{\linewidth}{!}{%
\begin{tabular}{lll}
\toprule
Scenario & Class signal $s_i(t)$ & Expected behaviour \\
\midrule
S1 Global & $z_i\{0.75\sin(2\pi t)+0.5\cos(4\pi t)\}$ & global signal \\
S2 LocalEarly & $1.3\,z_i\,\phi(t;0.16,0.035)$ & early transient \\
S3 LocalLate & $1.3\,z_i\,\phi(t;0.83,0.035)$ & late transient \\
S4 OppositePatches & $1.1\,z_i\{\phi(t;0.20,0.028)-\phi(t;0.77,0.028)\}$ & two opposite transients \\
S5 CrossWindowXOR & $1.45\{A_i\,\phi(t;0.19,0.04)+B_i\,\phi(t;0.82,0.04)\}$ & cross-window interaction \\
S6 Null & $0$ & no signal \\
S7 MixedLocalGlobal & $z_i\{0.38\sin(2\pi t)+0.8\,\phi(t;0.17,0.03)-0.7\,\phi(t;0.78,0.05)\}$ & mixed \\
S8 ImbalancedLocal & $0.85\,z_i\,\phi(t;0.20,0.06)$, $P(z_i=1)=0.18$ & imbalanced transient \\
S9 WeakTransient & $0.55\,z_i\,\phi(t;0.62,0.018)$ & weak, short transient \\
S10 LowVarianceLocal & $0.70\,z_i\,\phi(t;0.82,0.03)$ + nuisance on $[0,0.6]$ & signal in low-variance region \\
S11 PhaseShift & $\alpha_i\,\phi(t;c_i,0.04)$, $c_i=0.45+0.05\,\ind(z_i=1)+\delta_i$ & phase variation \\
S12 DistributedTransients & $0.5\,z_i\{\phi(t;0.15,0.02)-\phi(t;0.50,0.02)+\phi(t;0.85,0.02)\}$ & several weak transients \\
\bottomrule
\end{tabular}}
\caption{Class signals of the simulated scenarios in model \eqref{eq:dgp}. Further details are given in Section~\ref{sec:simdesign}.}
\label{tab:dgp}
\end{table}

\paragraph{\emph{Scenario 1, global signal.}}
The class signal lies in the span of the two nuisance functions, so that discrimination requires combining information from the whole domain and the signal is partly aligned with the largest source of nuisance variability.

\paragraph{\emph{Scenarios 2--4, short transients.}}
In Scenarios 2 and 3 the classes differ by a single Gaussian bump with standard deviation $0.035$, located early or late in the domain; in Scenario 4 the classes differ by two bumps of opposite sign. These scenarios were designed to favour local representations, since each bump lies essentially within an interval of width four standard deviations, that is, less than 15\% of the domain.

\paragraph{\emph{Scenario 5, cross-window interaction.}}
Two bumps with independent Rademacher amplitudes $A_i$ and $B_i$ are located in distant parts of the domain, and $Y_i$ is determined by $A_iB_i$ as in \eqref{eq:xor_structure}. The class means coincide and every window is marginally uninformative, so that local voting cannot succeed while a classifier with access to both parts of the domain can.

\paragraph{\emph{Scenario 6, null.}}
No class signal is present; this scenario assesses whether the aggregation assigns weight to uninformative experts.

\paragraph{\emph{Scenarios 7 and 8, mixed and imbalanced signals.}}
In these two scenarios $u_{1i}\sim N(0,1.3^2)$ and $u_{2i}\sim N(0,1)$. Scenario 7 combines a weak global component with two local bumps of different width and sign. Scenario 8 has a single moderately wide bump, an additional independent white noise with standard deviation $0.2$ and an imbalanced design with $P(z_i=1)=0.18$.

\paragraph{\emph{Scenario 9, weak short transient.}}
The classes differ by a bump of amplitude $0.55$ and standard deviation $0.018$ at $t=0.62$. With respect to Scenarios 2 and 3, the signal is weaker and shorter, so that a tuned global forest is not expected to separate the classes almost perfectly.

\paragraph{\emph{Scenario 10, signal in a low-variance region.}}
Additional nuisance variability
\[
\sum_{m=1}^{6}v_{im}\sin(m\pi t/0.6)\,\ind(t\leq0.6),
\]
with $v_{im}\sim N(0,1.5^2)$, is confined to the first 60\% of the domain, and the class signal is a bump at $t=0.82$. A global FPCA truncated at a moderate threshold is dominated by the high-variance region, which is the situation that motivates multiscale FPCA \citep{linzhu2019}.

\paragraph{\emph{Scenario 11, phase variation.}}
All curves contain a bump of standard deviation $0.04$ with amplitude $\alpha_i=1.2(1+0.15\xi_i)$, $\xi_i\sim N(0,1)$; its location is $0.45$ in one class and $0.50$ in the other, with a curve-specific jitter $\delta_i\sim N(0,0.015^2)$. The classes therefore differ in phase rather than in amplitude, a situation in which FPCA is known to be inefficient \citep{marron2015}.

\paragraph{\emph{Scenario 12, distributed weak transients.}}
The classes differ by three weak, short bumps with alternating signs spread over the domain, so that the discriminative information is local but not confined to a single window.

\subsection{Results}\label{sec:simresults}

\begin{figure}[!p]
\centering
\includegraphics[width=\linewidth]{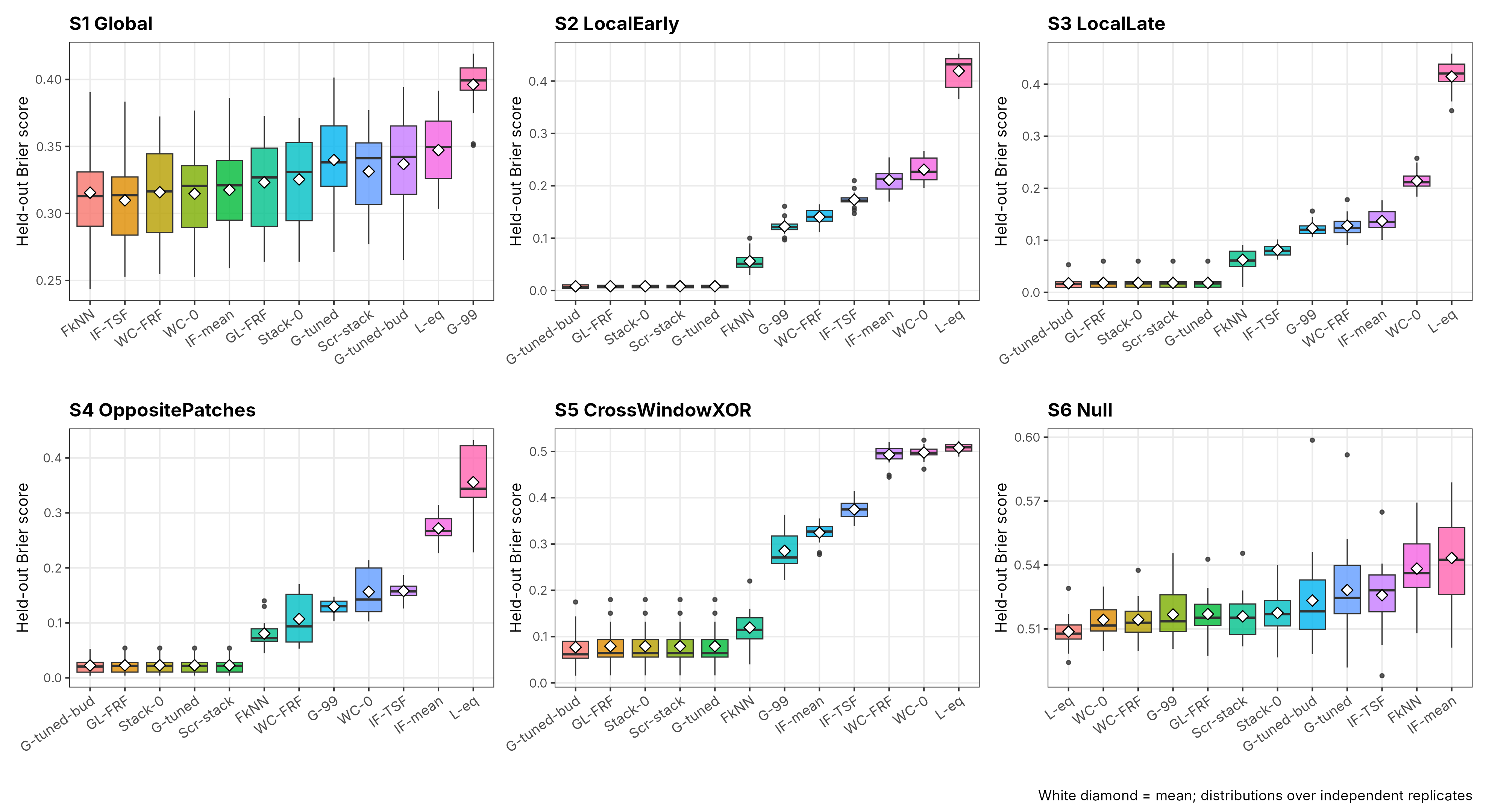}
\caption{Held-out Brier score of all methods in Scenarios 1--6 over 20 independent replicates. Methods are ordered by median; the white diamond indicates the mean. Vertical scales differ across panels.}
\label{fig:simbox_a}
\end{figure}

\begin{figure}[!p]
\centering
\includegraphics[width=\linewidth]{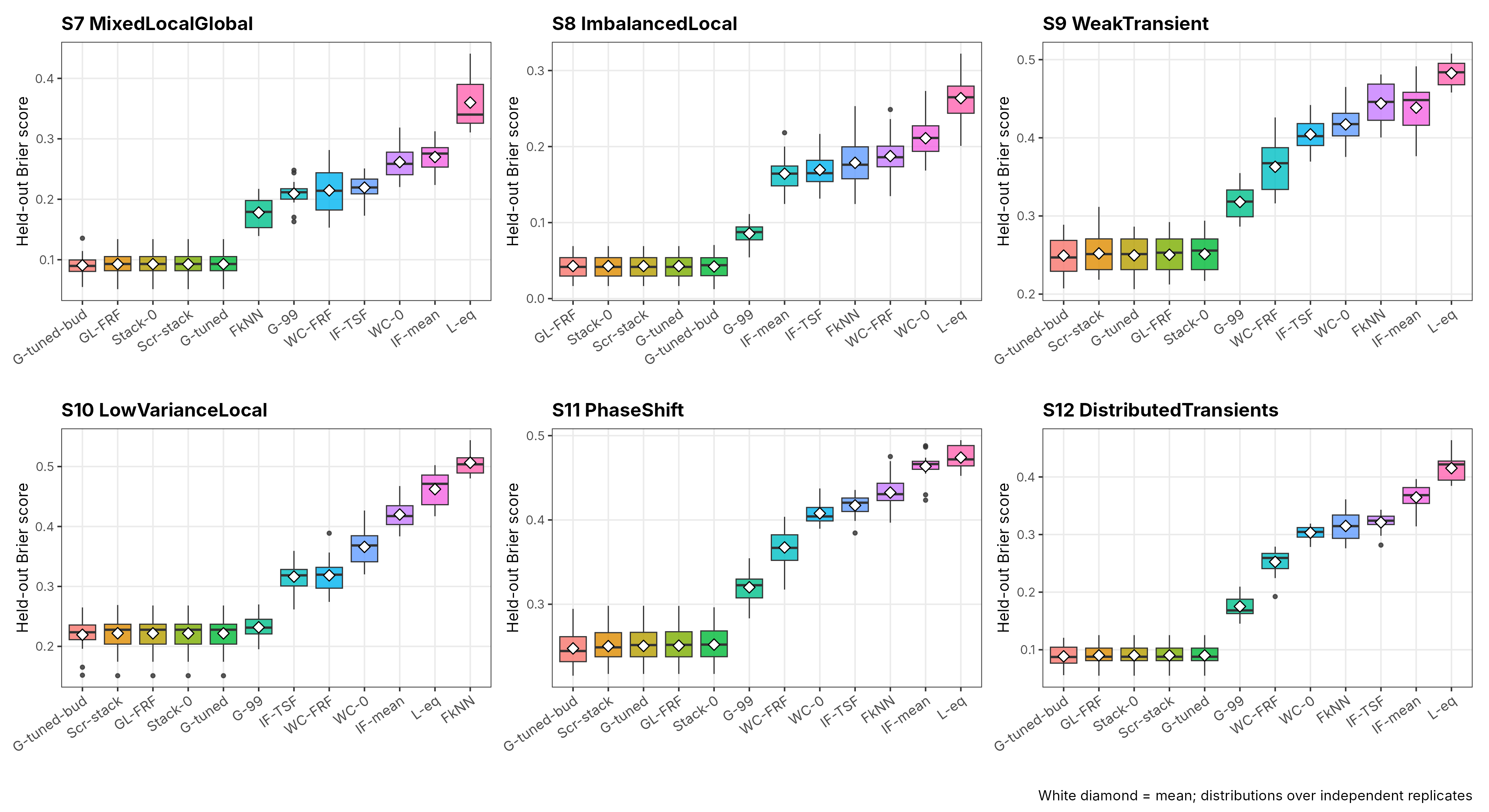}
\caption{Held-out Brier score of all methods in Scenarios 7--12 over 20 independent replicates. Methods are ordered by median; the white diamond indicates the mean. Vertical scales differ across panels.}
\label{fig:simbox_b}
\end{figure}

\begin{figure}[t]
\centering
\includegraphics[width=\linewidth]{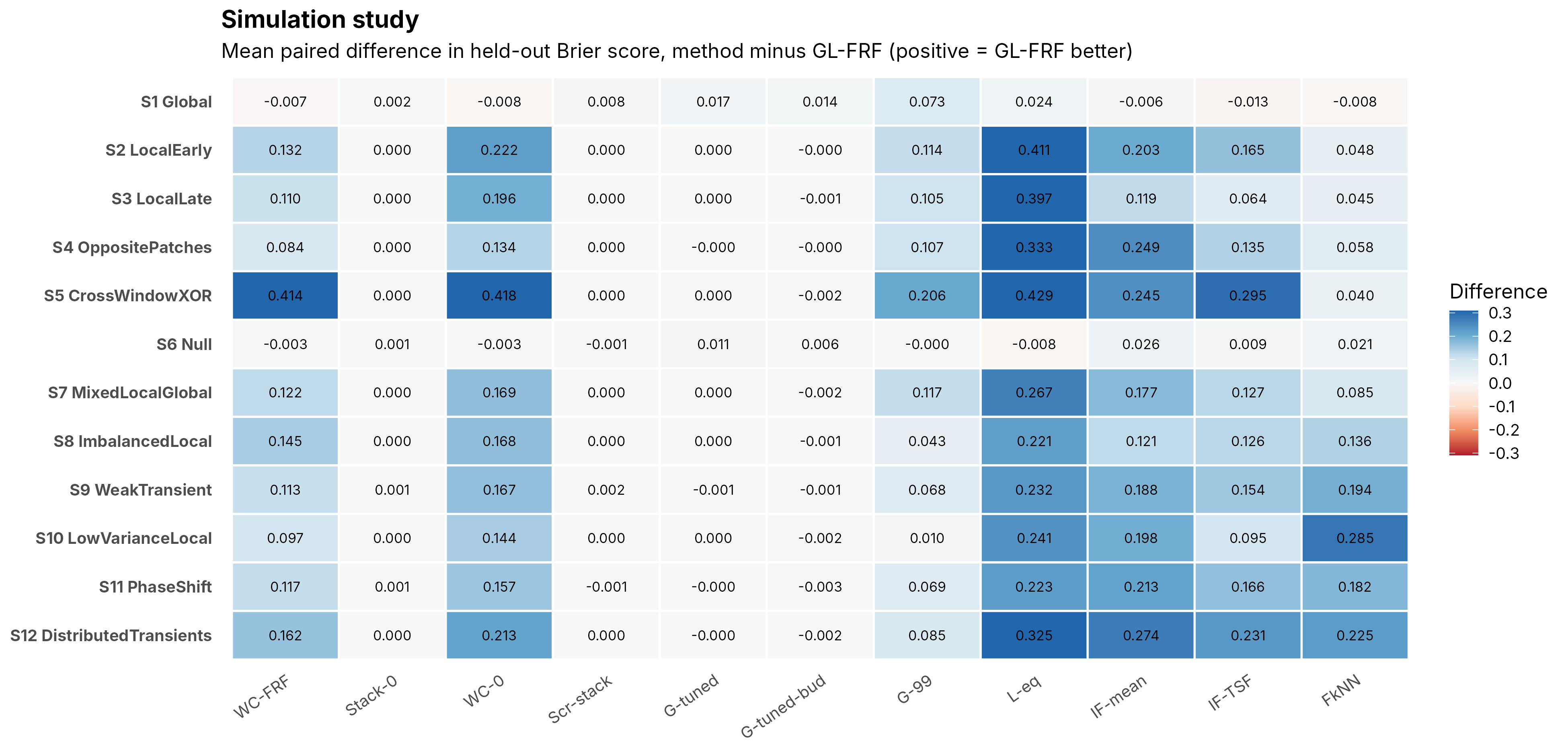}
\caption{Simulation study: mean paired difference in held-out Brier score between each competitor and GL-FRF. Blue cells indicate that GL-FRF has a lower Brier score, red cells that the competitor has a lower Brier score.}
\label{fig:heat_sim}
\end{figure}

Tables~\ref{tab:sim_brier} and~\ref{tab:sim_ba} report the mean Brier score and balanced accuracy of all methods, Table~\ref{tab:sim_paired} the paired differences with GL-FRF and their Monte Carlo standard errors, Figures~\ref{fig:simbox_a} and~\ref{fig:simbox_b} the distributions of the held-out Brier score, and Figure~\ref{fig:heat_sim} the mean paired differences for all competitors. The projected gradient algorithm converged in all fits, and no local smoothing parameter was selected on the boundary of its grid.

\paragraph{Tuning of the global expert.}
Internal cross-validation selected a threshold of 80\% or 90\% for the global expert in every replicate of the scenarios with a localised signal, except Scenario 10, retaining on average between 3.5 and 19.1 components, against 49 to 65 components at 99\%. In Scenario 10, in which the class signal lies in the low-variance part of the domain, the 99\% threshold was selected in 18 of 20 replicates. The effect of tuning is large: in the eleven scenarios with a class signal, the untuned global forest G-99 had a mean Brier score between 0.010 (Scenario 10) and 0.206 (Scenario 5) higher than the tuned global expert. In other words, a moderate truncation of the global FPCA, chosen by cross-validation, already separates short transients and cross-window interactions that a 99\% truncation dilutes.

\paragraph{Contribution of the local experts.}
In ten of the twelve scenarios the median weight of the global expert was at least 0.998 (Figure~\ref{fig:globalweight}), and $\widehat a_0\geq0.999$ in every replicate of Scenarios 2--5, 7, 8 and 10, in 95\% of the replicates of Scenario 12, in 65\% of those of Scenario 11 and in 45\% of those of Scenario 9. In these scenarios GL-FRF and G-tuned differed by less than 0.001 in mean Brier score, and G-tuned-bud, which uses $100(\widehat W+1)$ trees, was lower than GL-FRF by at most 0.0033 (Scenario 11), an effect of the larger number of trees. The local experts received substantial weight only in Scenario 1, with median $\widehat a_0=0.37$, where GL-FRF reduced the Brier score of G-tuned by 0.0167 (MCSE 0.0035) and that of G-tuned-bud by 0.0137 (0.0033), and in the null Scenario 6, with median $\widehat a_0=0.38$. In the latter case GL-FRF was lower than G-tuned by 0.011 (0.004), but its balanced accuracy remained at chance level, so the reduction reflects the averaging of uninformative probability vectors and not discriminative information.

Scenarios 9--12 were designed to create conditions in which the windows should help, namely a weak and short transient, a signal in a low-variance region, a phase shift and several distributed transients. They did not: the tuned global forest reached balanced accuracies of 0.85, 0.90, 0.83 and 0.95, respectively, and every method relying only on local or interval information was markedly worse, with a Brier score higher than GL-FRF by 0.10 to 0.16 for WC-FRF and by 0.09 to 0.23 for IF-TSF (Table~\ref{tab:sim_paired}). Within the range of signals considered, a smoothed global FPCA with a selected threshold represents a localised difference among a moderate number of components, and the window-specific experts add nothing that the stack can use.

\paragraph{Locality control.}
Since the local experts received almost no weight, GL-FRF and Scr-stack coincided in Scenarios 2--5, 7, 8 and 12 and differed by less than 0.002 in Scenarios 9--11. The only scenario in which the two stacks differed was Scenario 1, where the contiguous windows were better than the scrambled partitions by 0.0082 (0.0022). The simulation therefore provides no evidence on the locality question beyond this scenario, a point taken up with the real data.

\paragraph{WC-FRF and the other competitors.}
Methods based only on window-specific scores or intervals failed where expected and also where they were not expected to fail. Under the cross-window interaction of Scenario 5, WC-0 and WC-FRF had Brier scores of 0.498 and 0.493, close to the value of a constant prediction, against 0.079 for GL-FRF: although WC-FRF can in principle split on two windows within the same tree, this is consistent with the known difficulty of greedy recursive partitioning with pure interactions, in which no single variable offers an informative first split. The local equal vote was at chance level in Scenarios 5 and 8 and among the two worst methods in every scenario with a localised signal. In Scenarios 2--4 and 7--12, WC-FRF, IF-mean and IF-TSF were worse than GL-FRF by 0.06 to 0.28, because the tuned full-curve forest, which they do not include, was already close to the best method. The exception is Scenario 1, in which IF-TSF, WC-FRF and FkNN had lower Brier scores than GL-FRF, by 0.0134 (0.0034), 0.0074 (0.0025) and 0.0077 (0.0037). Tuning the forest of WC-FRF reduced the Brier score of WC-0 in ten of the twelve scenarios. The penalties of GL-FRF made almost no difference: the mean Brier scores of GL-FRF and Stack-0 differed by at most 0.0022, in Scenario 1.

\input{tables/Table_Sim_Brier.tex}
\input{tables/Table_Sim_BA.tex}
\input{tables/Table_Sim_Paired.tex}

\section{Application to benchmark datasets}\label{sec:real}

\subsection{Datasets and resampling}\label{sec:realdata}

In this section, we analyse twenty univariate datasets from the UCR time series archive \citep{dau2019}, available at \url{https://www.timeseriesclassification.com/}: ChlorineConcentration, Coffee, Computers, ECG200, Earthquakes, Fish, FordA, FreezerRegularTrain, GunPoint, Ham, Herring, Meat, PowerCons, ProximalPhalanxOutlineCorrect, Strawberry, ToeSegmentation1, Trace, Wine, WormsTwoClass and Yoga. They include spectrographic, image-outline, motion, sensor and device data, binary and multiclass problems with up to seven classes, series lengths between 80 and 900 and training sizes between 28 and 3,601 curves. Because the nested selection of Section~\ref{sec:selection} requires at least twelve training curves per class, only datasets satisfying this condition are considered.

\begin{figure}[!p]
\centering
\includegraphics[width=\linewidth]{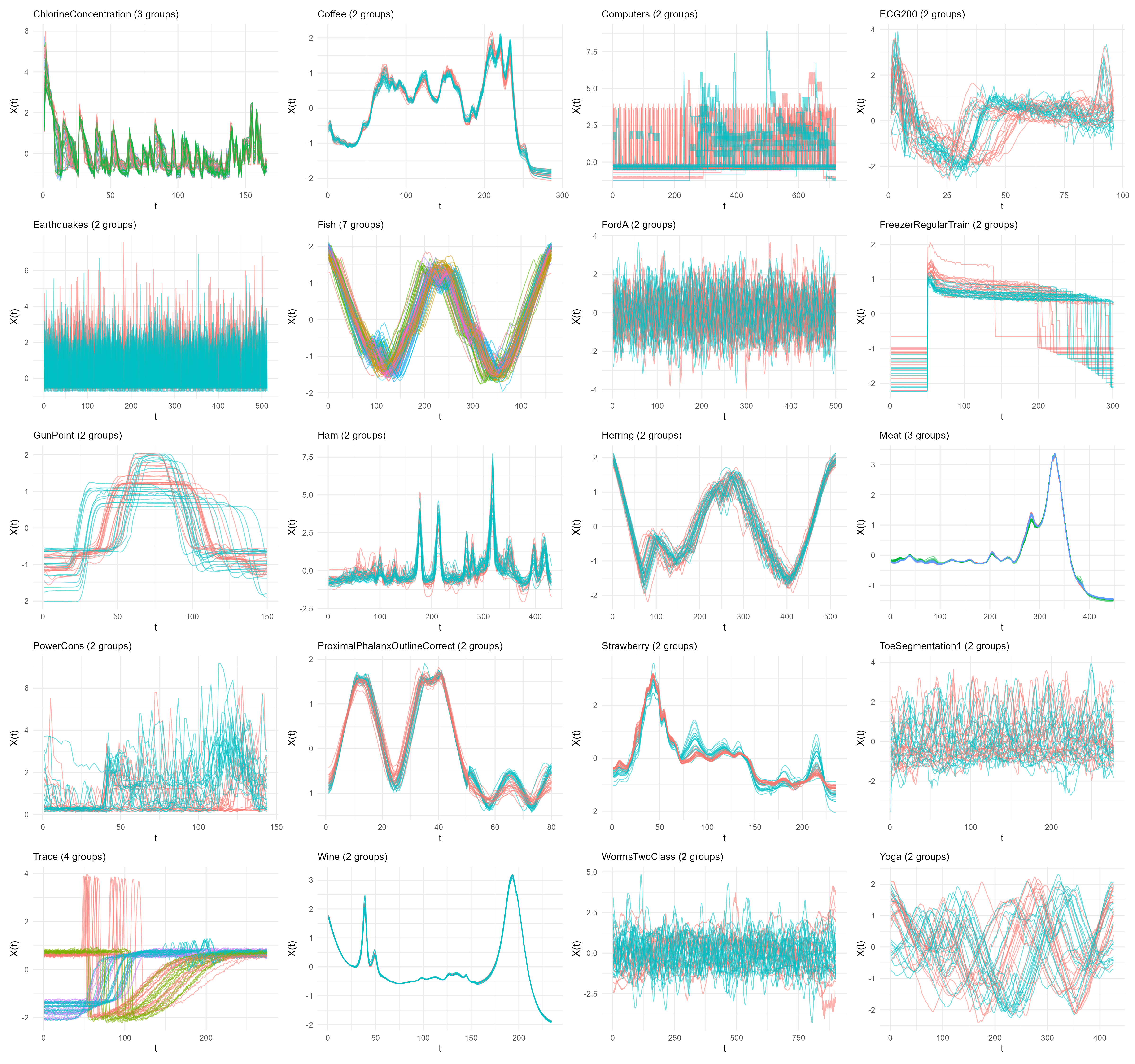}
\caption{Training curves of the twenty UCR datasets: up to 20 curves per class from the official training set, coloured by class. The number of classes is given in parentheses; $t$ indexes the points of the sampling grid and $X(t)$ is the value provided in the archive.}
\label{fig:ucrcurves}
\end{figure}

For each dataset, five training/test splits are analysed: the official UCR split and four stratified resamples, obtained by pooling the official training and test curves and drawing a training set with the official training class counts, as is customary in time series classification benchmarks \citep{bagnall2017}. FordA, which has 3,601 training curves, is analysed on its official split only for computational reasons. Table~\ref{tab:audit} reports the size of each dataset and the main quantities selected on the training sets. Series are complete and of equal length, and no amplitude standardisation is applied. Figure~\ref{fig:ucrcurves} displays training curves of all datasets.

\input{tables/Table_Data_Audit.tex}

\subsection{Statistical analysis}\label{sec:realanalysis}

For each dataset, method and split we compute the held-out Brier score and the paired difference with GL-FRF. Dataset-level performance is summarised by the mean over splits. Because resampled training sets overlap, the split-to-split variability of a paired difference is described by the corrected resampled interval of \citet{nadeaubengio2003}, which replaces the factor $1/R$ of the naive variance of the mean over $R$ splits with $1/R+n_{\mathrm{te}}/n_{\mathrm{tr}}$; these intervals are reported as descriptive measures.

For the family of main comparisons \{G-tuned-bud, Scr-stack, WC-FRF, IF-TSF\}, the dataset-level mean differences with GL-FRF are compared with zero by the two-sided Wilcoxon signed-rank test across the twenty datasets \citep{demsar2006}, and the four $p$-values are adjusted by the Holm procedure \citep{holm1979}. The family answers four questions: whether the window-specific representation improves on a tuned global forest with the same number of trees, whether its contribution is due to locality, whether late fusion differs from early fusion of the same representation, and whether GL-FRF is competitive with an established interval-based forest. Comparisons with the other methods are descriptive. All computations, including those of the simulation study, were carried out in R 4.5.2 under Windows, with forests fitted by the \texttt{ranger} package \citep{wright2017}, version 0.18.0; small numerical differences can arise with other versions of R and \texttt{ranger}.

\subsection{Results}\label{sec:ucrresults}

\begin{figure}[!p]
\centering
\includegraphics[width=\linewidth]{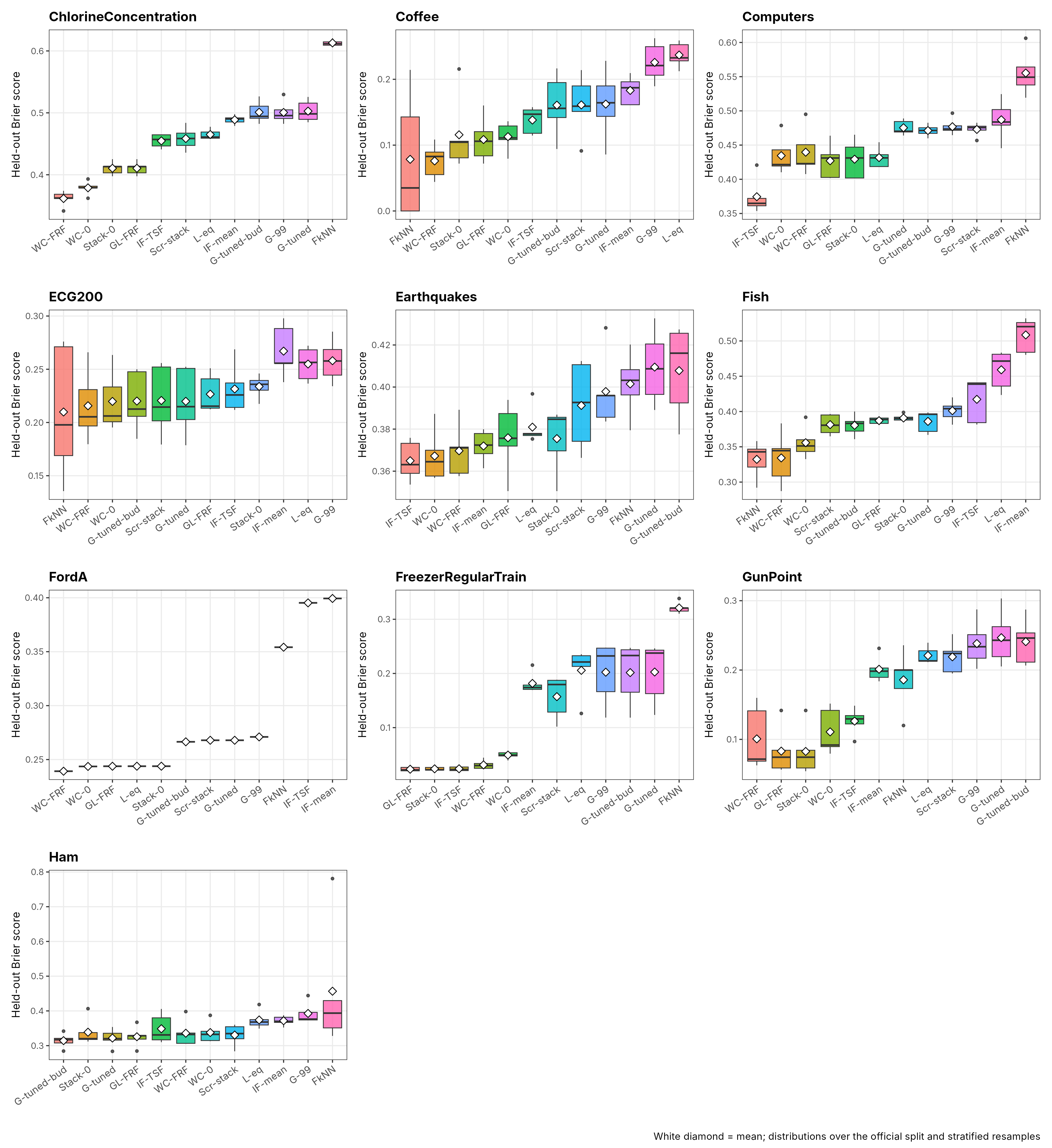}
\caption{UCR datasets (first ten in alphabetical order): held-out Brier score of all methods over the official split and four stratified resamples (FordA: official split only). Methods are ordered by median; the white diamond indicates the mean.}
\label{fig:ucrbox_a}
\end{figure}

\begin{figure}[!p]
\centering
\includegraphics[width=\linewidth]{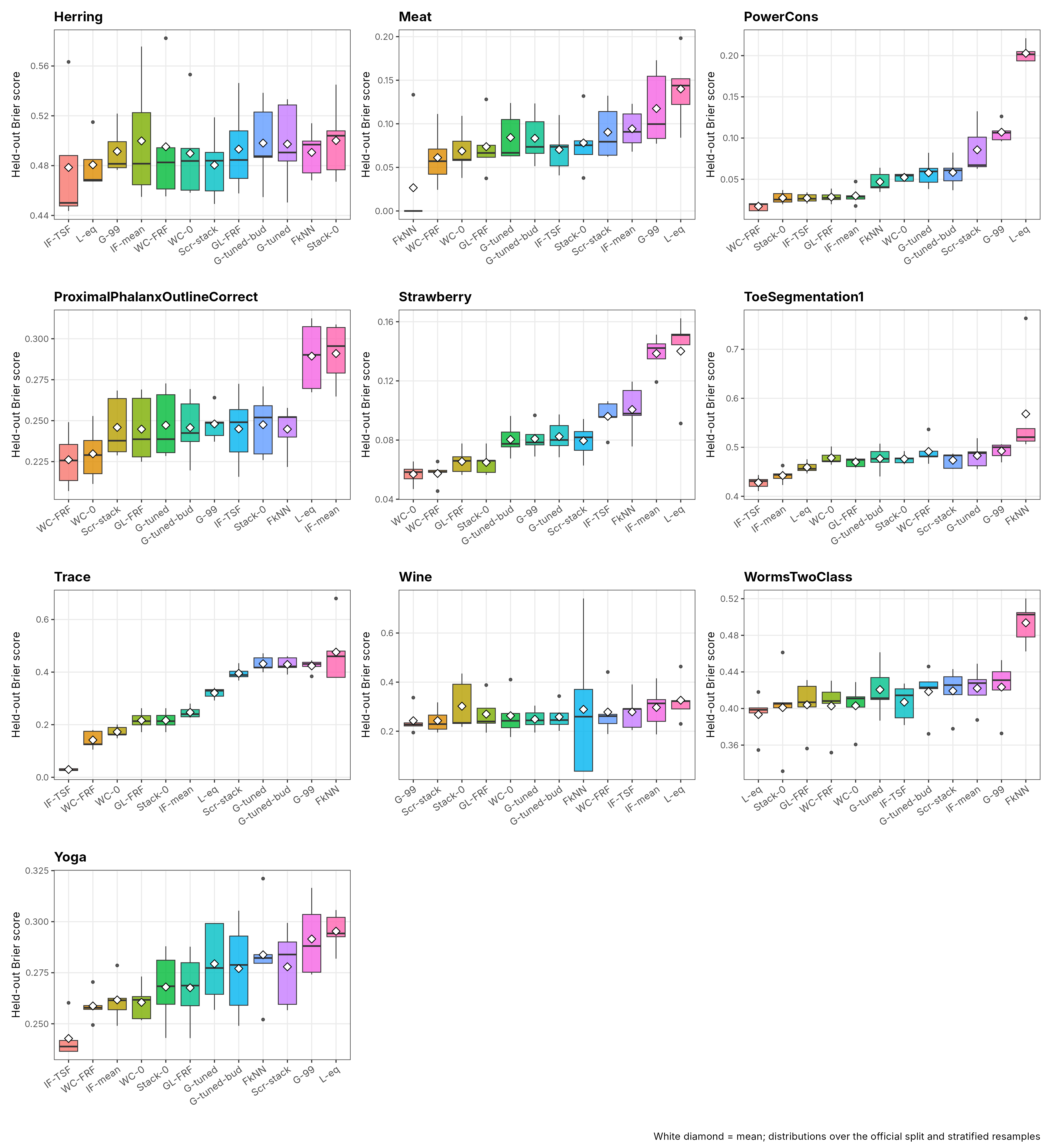}
\caption{UCR datasets (last ten in alphabetical order): held-out Brier score of all methods over the official split and four stratified resamples. Methods are ordered by median; the white diamond indicates the mean.}
\label{fig:ucrbox_b}
\end{figure}

\begin{figure}[t]
\centering
\includegraphics[width=\linewidth]{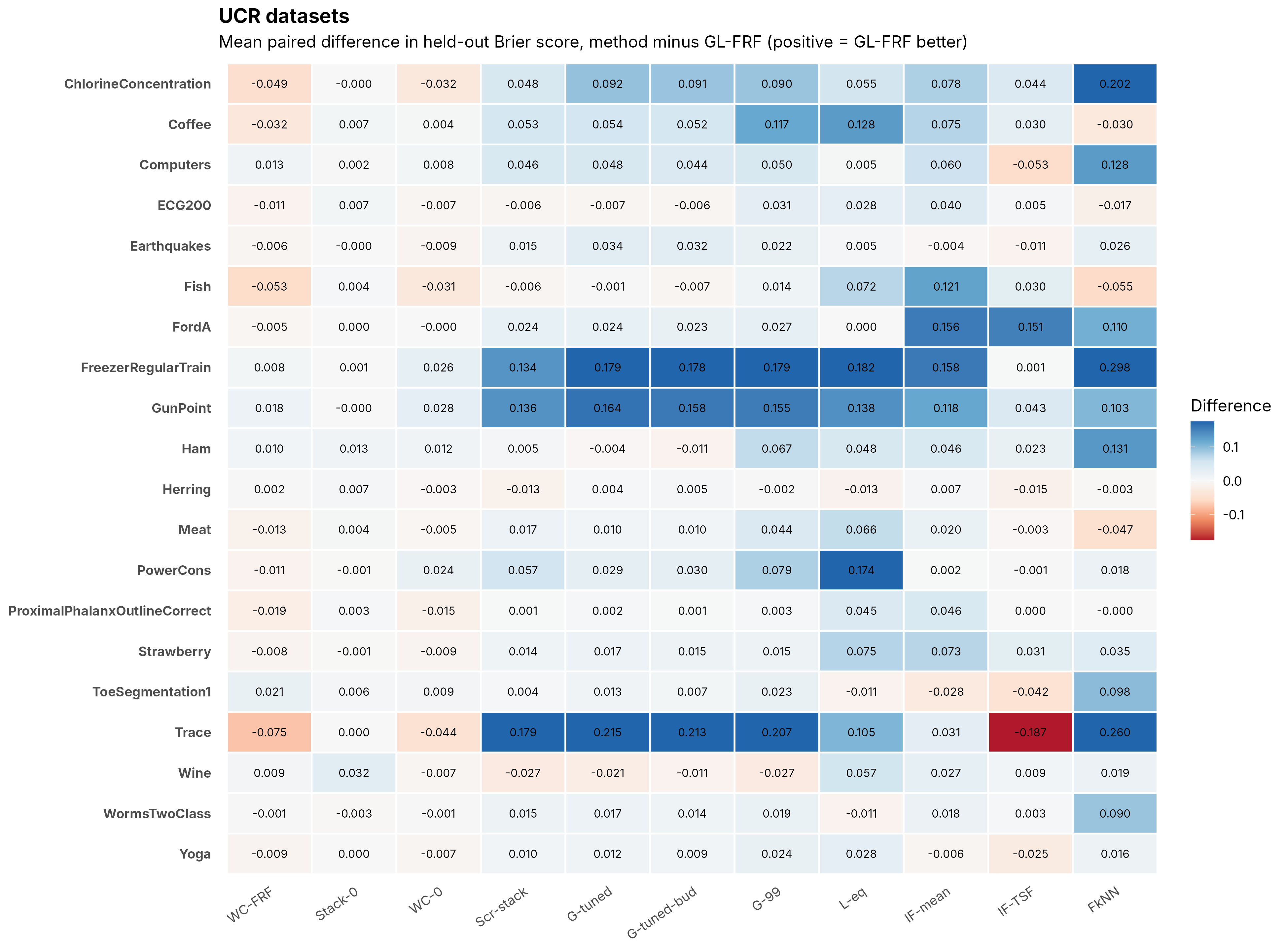}
\caption{UCR datasets: mean paired difference in held-out Brier score between each competitor and GL-FRF over the analysed splits. Blue cells indicate that GL-FRF has a lower Brier score.}
\label{fig:heat_real}
\end{figure}

\begin{figure}[t]
\centering
\includegraphics[width=.95\linewidth]{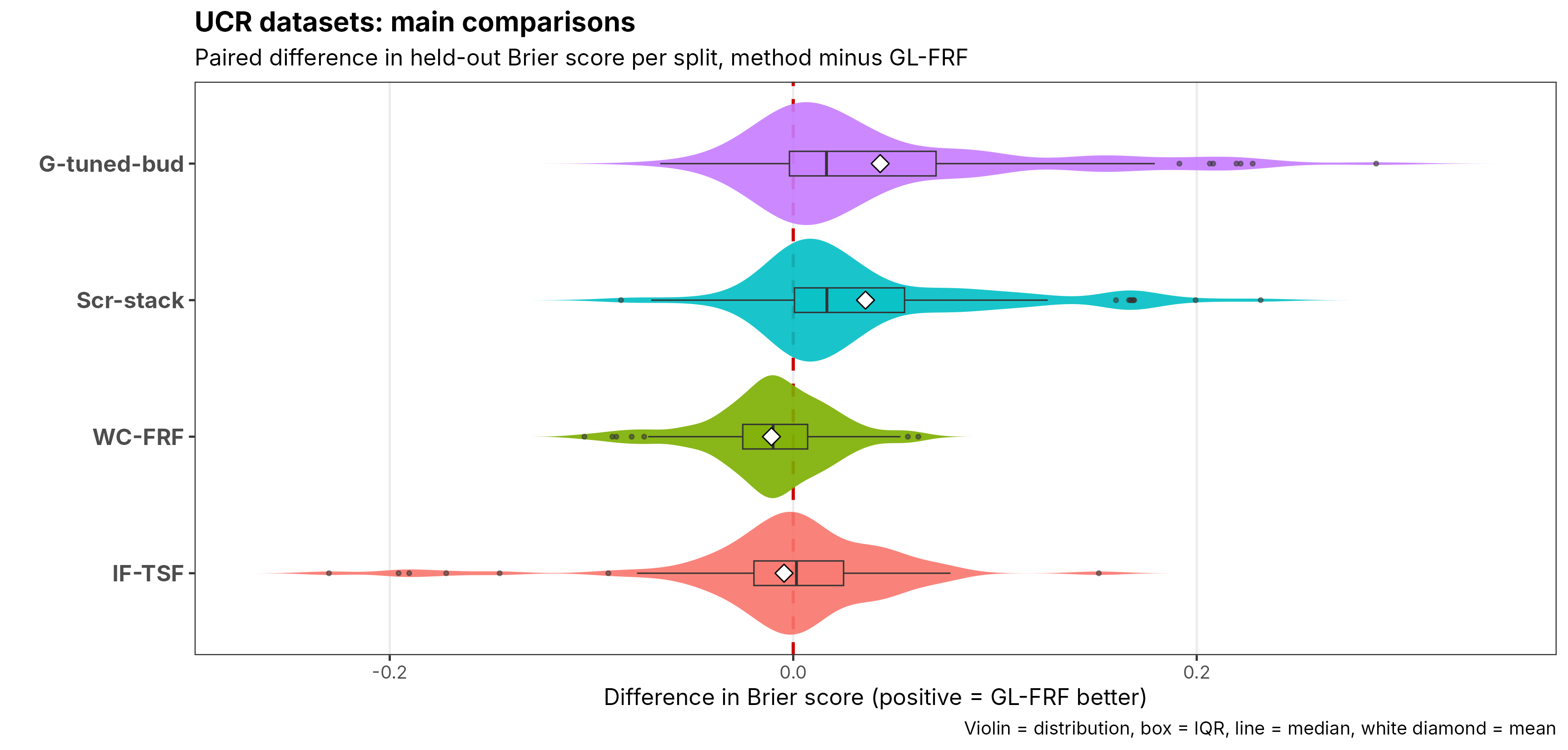}
\caption{UCR datasets: distribution over datasets and splits of the paired difference in held-out Brier score between each competitor of the main comparisons and GL-FRF. Positive values favour GL-FRF.}
\label{fig:maindiff}
\end{figure}

Tables~\ref{tab:ucr_brier} and~\ref{tab:ucr_ba} and Figures~\ref{fig:ucrbox_a}, \ref{fig:ucrbox_b} and~\ref{fig:heat_real} report the results on the twenty datasets, averaged over the official split and four stratified resamples, and Table~\ref{tab:ucr_tests} and Figure~\ref{fig:maindiff} the main comparisons. GL-FRF had a lower mean Brier score than the tuned global forest with the same number of trees on 16 of the 20 datasets, with a median difference of 0.015. The largest reductions were observed on Trace (0.213), FreezerRegularTrain (0.178), GunPoint (0.158), ChlorineConcentration (0.091), Coffee (0.052) and Computers (0.044); on FreezerRegularTrain, the balanced accuracy of GL-FRF was 0.992 against 0.859. G-tuned-bud was better on ECG200 (0.006), Fish (0.007), Ham (0.011) and Wine (0.011). The weight of the global expert was zero in every split of ChlorineConcentration, FordA, FreezerRegularTrain, GunPoint and Trace, and 0.12 and 0.40 on average on Computers and Coffee. Contrary to the simulation study, the window-specific experts therefore carried information that the tuned global decomposition did not represent.

The locality control indicates that this information is mainly due to the temporal localisation of the windows. Scr-stack, which has the same number of experts, the same set sizes and the same hyperparameters as GL-FRF, had a higher Brier score than GL-FRF on 16 datasets, with a median difference of 0.015, and recovered only a small part of the gain on the datasets where the windows helped: its Brier score was 0.395 on Trace, 0.157 on FreezerRegularTrain, 0.219 on GunPoint and 0.459 on ChlorineConcentration, against 0.216, 0.023, 0.083 and 0.410 for GL-FRF and 0.429, 0.202, 0.241 and 0.501 for G-tuned-bud. Diversity among forests trained on different subsets of the grid is therefore not sufficient to explain the improvement. Scr-stack was better than GL-FRF on Wine (0.027), Herring (0.013), ECG200 (0.006) and Fish (0.006).

After Holm adjustment, GL-FRF was significantly better than G-tuned-bud (unadjusted $p=0.001$, adjusted $p=0.006$) and than Scr-stack (unadjusted $p=0.003$, adjusted $p=0.009$). WC-FRF had a lower Brier score than GL-FRF on 13 datasets, with a median difference of 0.007 in its favour, but the difference was not significant ($p=0.143$, adjusted 0.286). GL-FRF and IF-TSF had a median difference of 0.002, with twelve datasets in favour of GL-FRF and eight against ($p=0.571$); IF-TSF was much better on Trace (0.030 against 0.216) and Computers (0.375 against 0.427) and much worse on FordA (0.395 against 0.244). No single method dominated: IF-TSF had the lowest mean Brier score on six datasets, WC-FRF on five, FkNN on three and each of the other methods on at most one. The corrected resampled intervals of the per-dataset differences were wide, because the test sets are large relative to the training sets on several datasets and the variance correction of \citet{nadeaubengio2003} is correspondingly large. Among the four main comparisons, they excluded zero in favour of GL-FRF for G-tuned-bud on ChlorineConcentration, GunPoint and Trace and for Scr-stack on Trace and WormsTwoClass, and in favour of the competitor for WC-FRF on ProximalPhalanxOutlineCorrect and Trace and for IF-TSF on Trace.

\input{tables/Table_UCR_Brier.tex}
\input{tables/Table_UCR_BA.tex}
\input{tables/Table_UCR_Tests.tex}

\subsection{GL-FRF and WC-FRF: when each method is preferable}\label{sec:glwc}

\begin{figure}[t]
\centering
\includegraphics[width=.85\linewidth]{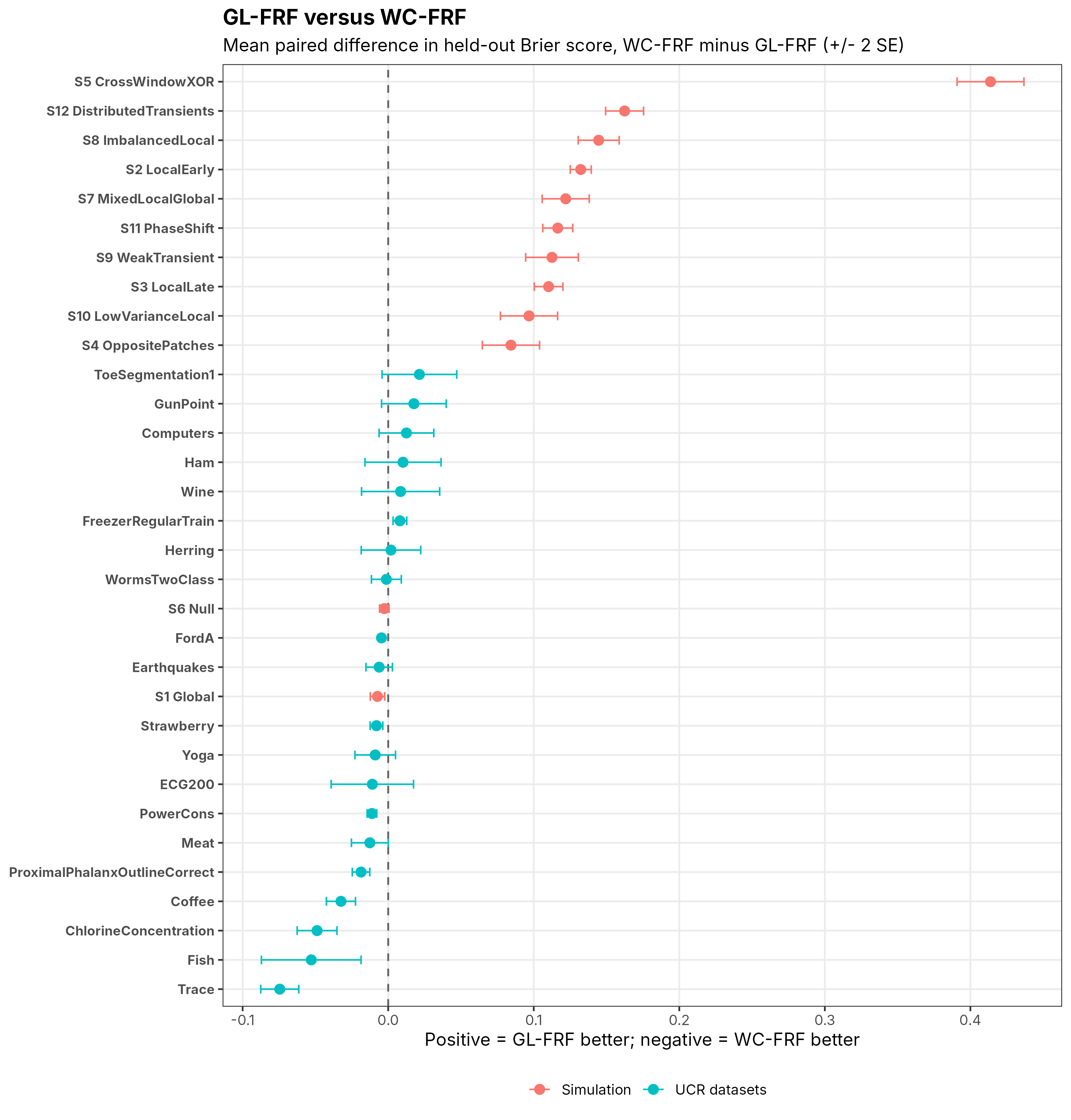}
\caption{Paired difference in held-out Brier score between WC-FRF and GL-FRF (mean $\pm$ two standard errors over replicates or splits). Positive values favour GL-FRF, negative values WC-FRF.}
\label{fig:glwc}
\end{figure}

Figure~\ref{fig:glwc} compares the two proposed methods in every simulated scenario and dataset. The comparison separates the settings into two groups. In ten of the twelve simulated scenarios GL-FRF was better, by 0.08 to 0.41, because the signal was captured by the tuned full-curve forest, which WC-FRF does not include; the largest difference occurred under the cross-window interaction of Scenario 5. WC-FRF was better in Scenario 1, by 0.007, and under the null scenario, by 0.003. On the real datasets the situation was reversed: WC-FRF was better on 13 of the 20 datasets, with the largest differences on Trace (0.075), Fish (0.053), ChlorineConcentration (0.049) and Coffee (0.032), and GL-FRF was better on seven, with the largest differences on ToeSegmentation1 (0.021), GunPoint (0.018) and Computers (0.013). The out-of-bag tuning of WC-FRF selected more candidate variables than the default $\lfloor\sqrt{K_W}\rfloor$ in 86\% of the fits, with a median close to $K_W/2$, and tuning reduced the Brier score of WC-0 on 14 of the 20 datasets.

The two methods therefore answer different needs. When the window-specific representation is known or expected to be informative, WC-FRF was usually the more accurate of the two methods. When the location and nature of the signal are unknown, GL-FRF is the safer choice, because it cannot be much worse than a tuned full-curve forest and it exploits the windows when they help. Table~\ref{tab:excess} quantifies this trade-off with the excess of each method over the best of the twelve methods in each setting. Across the 32 settings, GL-FRF had the smallest mean excess (0.021) and the smallest maximum excess (0.187, on Trace) of all methods. WC-FRF had a smaller mean excess than GL-FRF on the real datasets (0.021 against 0.031) but the largest maximum excess overall (0.416, in Scenario 5), and the global forests showed the opposite pattern, close to the best in the simulations and up to 0.399 from it on the real datasets. This summary is descriptive; it describes the behaviour of the methods across the settings considered and is not a test.

\input{tables/Table_Excess.tex}

\subsection{Selected functional representations and stacking weights}\label{sec:weights}

\begin{figure}[!p]
\centering
\includegraphics[width=.9\linewidth]{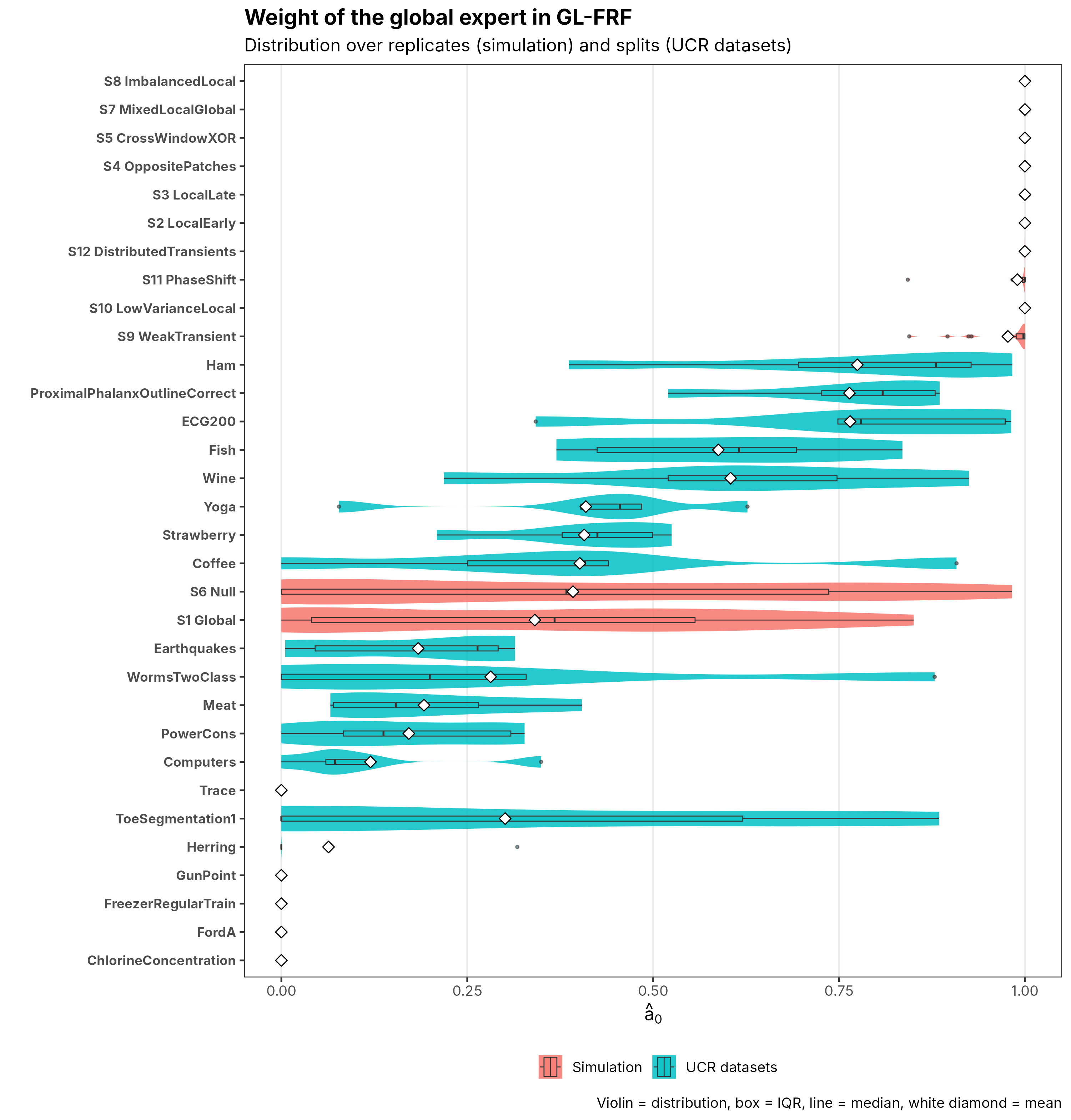}
\caption{Weight $\widehat a_0$ assigned to the tuned global expert by GL-FRF, over the replicates of each simulated scenario and the splits of each UCR dataset. Violin: distribution; box: interquartile range; line: median; white diamond: mean.}
\label{fig:globalweight}
\end{figure}

Figure~\ref{fig:globalweight} shows the weight of the global expert. In the simulations it was concentrated at one, except in Scenarios 1 and 6. On the real datasets it varied from zero in every split of ChlorineConcentration, FordA, GunPoint, Trace and FreezerRegularTrain, which are the datasets with the largest gains over the global forest, to a mean of 0.76 or more on ECG200, Ham and ProximalPhalanxOutlineCorrect, on which GL-FRF and the global forest performed similarly. The weight of the global expert is therefore informative about the usefulness of the windows on a given dataset.

\begin{figure}[!p]
\centering
\includegraphics[width=\linewidth]{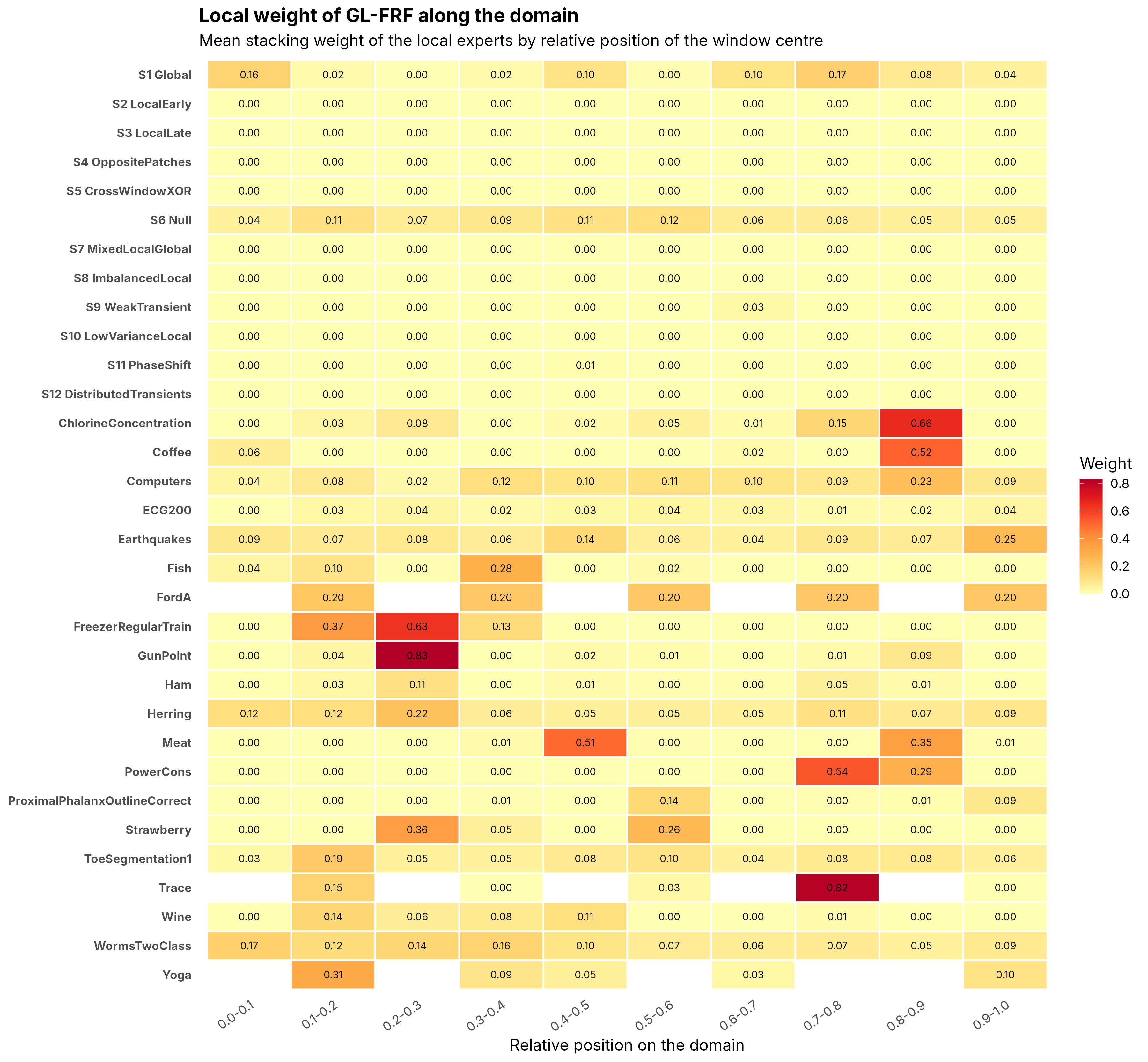}
\caption{Mean stacking weight assigned by GL-FRF to the local experts, by relative position of the window centre on the domain. For the simulated scenarios the location of the class signal is known (Table~\ref{tab:dgp}).}
\label{fig:localprofile}
\end{figure}

Figure~\ref{fig:localprofile} shows where the local weight was placed. On the datasets with large gains it was concentrated on one or two windows: 83\% of the weight on the window centred between 0.2 and 0.3 of the domain for GunPoint, 82\% between 0.7 and 0.8 for Trace, 63\% and 37\% on two adjacent windows in the first third of the domain for FreezerRegularTrain, and 66\% between 0.8 and 0.9 for ChlorineConcentration. On datasets with small or no gains the weight was spread over the domain. These profiles indicate the parts of the curves on which the local representation was useful; they describe the predictive contribution of the windows and are not evidence of a causal role of the corresponding parts of the signal.

The number of windows varied across splits on most datasets, and five, ten and twenty windows were all selected. At least one of the two penalties was positive in most fits; GL-FRF had a lower Brier score than Stack-0 on 14 of the 20 datasets, but the differences were small (median 0.002, maximum 0.032 on Wine), so the penalties had a small effect on accuracy. The median computing time of GL-FRF was between 37 seconds (ECG200) and 451 seconds (WormsTwoClass) per split, and 2,422 seconds on FordA, on a single core of a desktop computer; the whole evaluation, including all competitors, required about 90 minutes on 11 cores.

\section{Discussion and conclusions}\label{sec:discussion}

This paper introduced window-specific functional principal components as a representation for functional random forests and proposed two ways of exploiting it: GL-FRF, which stacks a tuned full-curve forest and one forest per window with penalised convex weights, and WC-FRF, which trains one tuned forest on the concatenated window-specific scores. The evaluation asked three questions: whether the window-specific representation adds information to a tuned full-curve forest, whether any gain is due to temporal localisation or to the diversity of an ensemble, and which of the two aggregation strategies should be preferred, and when.

On the first question, the simulation study and the benchmark datasets gave different answers. In the twelve simulated scenarios, which include short transients, a signal in a region of low variance, a phase shift, distributed transients and a cross-window interaction, a full-curve forest whose FPCA threshold and node size were chosen by internal cross-validation was already close to the best method, and in ten of them GL-FRF reduced to it almost exactly by assigning it all the weight. Simulated mechanisms built to favour local representations can thus be solved by a well-tuned global representation, and comparisons with untuned global forests overstate the value of windows. The simulations also show that GL-FRF pays essentially no price when the windows are not needed, since it then reduces to the tuned full-curve forest. On the benchmark datasets, instead, the window-specific representation improved on the tuned full-curve forest with the same number of trees on 16 of the 20 datasets, with reductions in the Brier score of up to 0.21 and a difference that remained significant after multiplicity adjustment; the largest gains were concentrated on a few datasets, while on the others the differences were small.

On the second question, the locality control recovered little of the gain on the datasets where the windows helped, which indicates that the improvement is mainly due to the temporal localisation of the windows rather than to the diversity of the ensemble, although the two stacks also differ in the smoothing of the curves. On those datasets the stacking weights of GL-FRF were concentrated on one or two windows, which provides a direct indication of the parts of the domain that contribute to the prediction.

On the third question, the two methods are complementary rather than competing. WC-FRF was the more accurate on real data, with a smaller mean excess over the best method than GL-FRF and a lower Brier score on 13 of 20 datasets, although the difference was not significant. It has no access to the full-curve representation, however, and was markedly worse than GL-FRF in all simulated scenarios in which the tuned full-curve forest captured the signal, and close to a constant prediction under the cross-window interaction. GL-FRF was rarely the single best method, but no method was best on more than six of the twenty datasets, and GL-FRF had the smallest mean excess over the best method (0.021) and the smallest maximum excess (0.187) across all 32 settings: its worst result was closer to the best method than the worst result of any other method. In practice, we suggest WC-FRF when the discriminative information is expected to be local and the curves are long enough to support several windows, and GL-FRF when the location and nature of the signal are unknown, when an interpretable decomposition of the prediction into global and local contributions is useful, or when a single method is needed whose accuracy does not depend on knowing in advance where the signal lies. The weight of the global expert estimated by GL-FRF can itself be used as a diagnostic: values close to zero indicate that the window-specific representation is informative, and that WC-FRF is likely to be competitive. The penalties of GL-FRF had a minor role: GL-FRF behaves essentially as a cross-fitted convex stack with exact zero weights, and the penalties make the weights unique (Property 4); whether they reduce the variability of the weights on small training sets was not assessed.

The study has several limitations. The number of candidate variables at each split was tuned only for WC-FRF, the local experts used a fixed 99\% threshold, and the windows have equal width. WC-FRF and several competitors, including the locality control, inherit the number of windows selected for GL-FRF, which is not equivalent to tuning each of them separately, and WC-FRF does not include the full-curve scores. The untuned concatenated forest WC-0 follows the approach of \citet{linzhu2019} with equal-width windows, but a partition based on the variance function, as in multiscale FPCA, was not evaluated. The interval forests are implementations of the time series forest principle and not the full Canonical Interval Forest, and state-of-the-art time series classifiers were not included. The twenty datasets are a small and non-random sample of the UCR archive, so that the results may not extend to other collections of curves. Finally, the simulated scenarios did not reproduce the situations in which the windows were useful on real data, so the conditions under which a window-specific representation outperforms a tuned global one remain to be characterised.

These results suggest four directions for future research: a hybrid that adds the full-curve scores to the concatenated window-specific scores of WC-FRF, which would combine the strengths of the two methods; data-driven or overlapping windows instead of an equal-width partition, possibly informed by localised or multiscale FPCA \citep{chenlei2015,linzhu2019}; registration of the curves before the decomposition when classes differ in phase \citep{marron2015}; and explainability tools for both methods, extending those developed for functional random forests \citep{maturoadac2026} to window-specific components.

\section*{Funding}
No funds, grants, or other support were received for the research reported in this article.

\section*{Conflict of interest}
The authors have no competing interests to declare that are relevant to the content of this article.

\section*{Use of generative AI}
The authors used generative AI tools to improve the English language, to assist in  revising parts of the text, and to check bibliographic metadata. All content was critically reviewed, verified and edited by the authors, who take full responsibility for the publication.

\section*{Data availability statement}
All datasets analysed in this study are publicly available from the UCR time series classification archive (\url{https://www.timeseriesclassification.com/}). The R code that implements the methods and reproduces all analyses, tables and figures, including the generation of the simulated data, is available from the corresponding author on request.

\section*{Author contributions}
Both authors contributed equally to all aspects of the article, including conceptualisation, methodology, software, formal analysis, writing and revision, and approved the final manuscript.


\end{document}

%% file: tables/Table_Sim_Brier.tex
\begin{sidewaystable}[p]
\centering
\scriptsize
\setlength{\tabcolsep}{2.5pt}
\resizebox{\linewidth}{!}{\begin{tabular}{lrrrrrrrrrrrr}
\toprule
Scenario & GL-FRF & WC-FRF & Stack-0 & WC-0 & Scr-stack & G-tuned & G-tuned-bud & G-99 & L-eq & IF-mean & IF-TSF & FkNN
\\
\midrule
S1 Global & 0.323 (0.034) & 0.316 (0.036) & 0.325 (0.034) & 0.315 (0.035) & 0.331 (0.032) & 0.340 (0.038) & 0.337 (0.038) & 0.396 (0.019) & 0.347 (0.026) & 0.318 (0.037) & \textbf{0.310 (0.037)} & 0.315 (0.038) \\
S2 LocalEarly & \textbf{0.008 (0.003)} & 0.141 (0.015) & \textbf{0.008 (0.003)} & 0.231 (0.023) & \textbf{0.008 (0.003)} & \textbf{0.008 (0.003)} & \textbf{0.008 (0.004)} & 0.122 (0.015) & 0.419 (0.029) & 0.211 (0.025) & 0.173 (0.014) & 0.056 (0.017) \\
S3 LocalLate & 0.018 (0.012) & 0.128 (0.020) & 0.018 (0.012) & 0.214 (0.019) & 0.018 (0.012) & 0.018 (0.012) & \textbf{0.017 (0.011)} & 0.123 (0.013) & 0.415 (0.031) & 0.137 (0.022) & 0.082 (0.012) & 0.063 (0.021) \\
S4 OppositePatches & 0.023 (0.014) & 0.107 (0.046) & 0.023 (0.014) & 0.157 (0.042) & 0.023 (0.014) & 0.023 (0.014) & \textbf{0.022 (0.014)} & 0.129 (0.012) & 0.356 (0.074) & 0.272 (0.025) & 0.158 (0.016) & 0.081 (0.023) \\
S5 CrossWindowXOR & 0.079 (0.040) & 0.493 (0.020) & 0.079 (0.040) & 0.498 (0.014) & 0.079 (0.040) & 0.079 (0.040) & \textbf{0.077 (0.038)} & 0.285 (0.044) & 0.508 (0.009) & 0.325 (0.020) & 0.375 (0.023) & 0.119 (0.038) \\
S6 Null & 0.517 (0.010) & 0.514 (0.009) & 0.517 (0.010) & 0.514 (0.008) & 0.516 (0.011) & 0.528 (0.021) & 0.523 (0.023) & 0.517 (0.011) & \textbf{0.509 (0.008)} & 0.543 (0.020) & 0.526 (0.017) & 0.538 (0.015) \\
S7 MixedLocalGlobal & 0.093 (0.019) & 0.215 (0.039) & 0.093 (0.019) & 0.262 (0.028) & 0.093 (0.019) & 0.093 (0.019) & \textbf{0.091 (0.019)} & 0.209 (0.020) & 0.360 (0.042) & 0.270 (0.026) & 0.219 (0.020) & 0.178 (0.026) \\
S8 ImbalancedLocal & 0.043 (0.015) & 0.187 (0.028) & 0.043 (0.015) & 0.211 (0.026) & 0.043 (0.015) & 0.043 (0.015) & \textbf{0.042 (0.015)} & 0.086 (0.014) & 0.264 (0.032) & 0.164 (0.024) & 0.169 (0.023) & 0.179 (0.035) \\
S9 WeakTransient & 0.250 (0.024) & 0.363 (0.031) & 0.251 (0.024) & 0.417 (0.023) & 0.252 (0.025) & 0.250 (0.024) & \textbf{0.249 (0.023)} & 0.318 (0.022) & 0.483 (0.016) & 0.439 (0.029) & 0.404 (0.019) & 0.444 (0.026) \\
S10 LowVarianceLocal & 0.222 (0.029) & 0.318 (0.030) & 0.222 (0.029) & 0.366 (0.028) & 0.222 (0.029) & 0.222 (0.029) & \textbf{0.219 (0.027)} & 0.232 (0.018) & 0.462 (0.028) & 0.420 (0.024) & 0.316 (0.024) & 0.506 (0.020) \\
S11 PhaseShift & 0.251 (0.020) & 0.367 (0.022) & 0.252 (0.019) & 0.408 (0.014) & 0.250 (0.019) & 0.251 (0.019) & \textbf{0.248 (0.020)} & 0.320 (0.018) & 0.474 (0.014) & 0.464 (0.015) & 0.417 (0.013) & 0.432 (0.020) \\
S12 DistributedTransients & 0.090 (0.020) & 0.253 (0.021) & 0.090 (0.020) & 0.303 (0.011) & 0.090 (0.020) & 0.090 (0.020) & \textbf{0.089 (0.020)} & 0.175 (0.020) & 0.415 (0.023) & 0.365 (0.021) & 0.321 (0.016) & 0.315 (0.025) \\
\bottomrule
\end{tabular}}
\caption{Simulation study: mean (SD) held-out multiclass Brier score over 20 replicates (lower is better). GL-FRF: global--local functional random forest (late fusion); WC-FRF: window-concatenated functional random forest (early fusion, OOB-tuned); Stack-0: GL-FRF without penalties; WC-0: WC-FRF with default hyperparameters; Scr-stack: the same stack, with the $(\widehat W,\widehat\lambda_G,\widehat\lambda_L)$ of GL-FRF, on random non-contiguous partitions of the grid (locality control); G-tuned: tuned global expert alone (100 trees); G-tuned-bud: the same with $100(\widehat W+1)$ trees; G-99: untuned global forest (99\% FPCA); L-eq: equal soft vote of the local experts; IF-mean, IF-TSF: random-interval forests with interval means, and with mean, standard deviation and slope; FkNN: functional $k$-nearest neighbours. Best mean in bold, including methods tied with it at the reported precision.}
\label{tab:sim_brier}
\end{sidewaystable}

%% file: tables/Table_Sim_BA.tex
\begin{sidewaystable}[p]
\centering
\scriptsize
\setlength{\tabcolsep}{2.5pt}
\resizebox{\linewidth}{!}{\begin{tabular}{lrrrrrrrrrrrr}
\toprule
Scenario & GL-FRF & WC-FRF & Stack-0 & WC-0 & Scr-stack & G-tuned & G-tuned-bud & G-99 & L-eq & IF-mean & IF-TSF & FkNN
\\
\midrule
S1 Global & 0.776 (0.033) & \textbf{0.778 (0.035)} & 0.770 (0.030) & \textbf{0.778 (0.034)} & 0.769 (0.042) & 0.764 (0.037) & 0.766 (0.035) & 0.751 (0.035) & 0.774 (0.036) & 0.770 (0.031) & \textbf{0.778 (0.034)} & 0.777 (0.037) \\
S2 LocalEarly & \textbf{0.998 (0.003)} & 0.929 (0.022) & \textbf{0.998 (0.003)} & 0.924 (0.024) & \textbf{0.998 (0.003)} & \textbf{0.998 (0.003)} & \textbf{0.998 (0.004)} & \textbf{0.998 (0.004)} & 0.794 (0.061) & 0.901 (0.032) & 0.965 (0.019) & 0.972 (0.012) \\
S3 LocalLate & \textbf{0.993 (0.008)} & 0.932 (0.022) & \textbf{0.993 (0.008)} & 0.938 (0.018) & \textbf{0.993 (0.008)} & \textbf{0.993 (0.008)} & \textbf{0.993 (0.007)} & 0.989 (0.009) & 0.805 (0.059) & 0.942 (0.018) & 0.989 (0.008) & 0.967 (0.014) \\
S4 OppositePatches & 0.993 (0.008) & 0.950 (0.030) & 0.993 (0.008) & 0.967 (0.023) & 0.993 (0.008) & 0.993 (0.008) & \textbf{0.994 (0.007)} & 0.990 (0.009) & 0.891 (0.085) & 0.865 (0.036) & 0.965 (0.019) & 0.957 (0.012) \\
S5 CrossWindowXOR & 0.960 (0.024) & 0.549 (0.056) & 0.960 (0.024) & 0.539 (0.038) & 0.960 (0.024) & 0.960 (0.024) & \textbf{0.963 (0.023)} & 0.897 (0.053) & 0.499 (0.037) & 0.811 (0.035) & 0.785 (0.046) & 0.933 (0.026) \\
S6 Null & 0.484 (0.029) & \textbf{0.502 (0.027)} & 0.492 (0.029) & 0.491 (0.031) & 0.492 (0.030) & 0.484 (0.036) & 0.490 (0.028) & 0.499 (0.034) & 0.497 (0.028) & 0.483 (0.035) & 0.485 (0.029) & 0.468 (0.023) \\
S7 MixedLocalGlobal & \textbf{0.964 (0.017)} & 0.869 (0.042) & \textbf{0.964 (0.017)} & 0.861 (0.051) & \textbf{0.964 (0.017)} & \textbf{0.964 (0.017)} & 0.962 (0.018) & 0.948 (0.017) & 0.793 (0.066) & 0.814 (0.034) & 0.867 (0.034) & 0.890 (0.024) \\
S8 ImbalancedLocal & \textbf{0.959 (0.026)} & 0.689 (0.069) & \textbf{0.959 (0.026)} & 0.583 (0.067) & \textbf{0.959 (0.026)} & \textbf{0.959 (0.026)} & 0.958 (0.026) & 0.942 (0.034) & 0.500 (0.000) & 0.691 (0.058) & 0.651 (0.055) & 0.696 (0.074) \\
S9 WeakTransient & 0.846 (0.028) & 0.752 (0.036) & 0.845 (0.028) & 0.724 (0.050) & \textbf{0.847 (0.026)} & \textbf{0.847 (0.027)} & 0.845 (0.024) & 0.842 (0.024) & 0.597 (0.054) & 0.643 (0.036) & 0.719 (0.032) & 0.644 (0.040) \\
S10 LowVarianceLocal & \textbf{0.900 (0.019)} & 0.789 (0.035) & \textbf{0.900 (0.019)} & 0.782 (0.042) & \textbf{0.900 (0.019)} & \textbf{0.900 (0.019)} & 0.897 (0.020) & 0.894 (0.019) & 0.649 (0.081) & 0.690 (0.042) & 0.842 (0.027) & 0.550 (0.039) \\
S11 PhaseShift & 0.832 (0.026) & 0.742 (0.030) & 0.834 (0.028) & 0.732 (0.034) & 0.832 (0.026) & 0.832 (0.026) & \textbf{0.835 (0.024)} & \textbf{0.835 (0.022)} & 0.620 (0.046) & 0.614 (0.029) & 0.709 (0.027) & 0.657 (0.025) \\
S12 DistributedTransients & \textbf{0.954 (0.016)} & 0.850 (0.023) & \textbf{0.954 (0.016)} & 0.873 (0.025) & \textbf{0.954 (0.016)} & \textbf{0.954 (0.016)} & \textbf{0.954 (0.014)} & 0.951 (0.015) & 0.784 (0.056) & 0.751 (0.037) & 0.819 (0.028) & 0.784 (0.028) \\
\bottomrule
\end{tabular}}
\caption{Simulation study: mean (SD) held-out balanced accuracy over 20 replicates (higher is better). GL-FRF: global--local functional random forest (late fusion); WC-FRF: window-concatenated functional random forest (early fusion, OOB-tuned); Stack-0: GL-FRF without penalties; WC-0: WC-FRF with default hyperparameters; Scr-stack: the same stack, with the $(\widehat W,\widehat\lambda_G,\widehat\lambda_L)$ of GL-FRF, on random non-contiguous partitions of the grid (locality control); G-tuned: tuned global expert alone (100 trees); G-tuned-bud: the same with $100(\widehat W+1)$ trees; G-99: untuned global forest (99\% FPCA); L-eq: equal soft vote of the local experts; IF-mean, IF-TSF: random-interval forests with interval means, and with mean, standard deviation and slope; FkNN: functional $k$-nearest neighbours. Best mean in bold, including methods tied with it at the reported precision.}
\label{tab:sim_ba}
\end{sidewaystable}

%% file: tables/Table_Sim_Paired.tex
\begin{sidewaystable}[p]
\centering
\scriptsize
\resizebox{\linewidth}{!}{\begin{tabular}{lrrrrrrr}
\toprule
Scenario & WC-FRF & G-tuned & G-tuned-bud & Stack-0 & Scr-stack & IF-TSF & FkNN
\\
\midrule
S1 Global & -0.0074 (0.0025) & 0.0167 (0.0035) & 0.0137 (0.0033) & 0.0022 (0.0011) & 0.0082 (0.0022) & -0.0134 (0.0034) & -0.0077 (0.0037) \\
S2 LocalEarly & 0.1323 (0.0036) & 0.0000 (0.0000) & -0.0001 (0.0003) & 0.0000 (0.0000) & 0.0000 (0.0000) & 0.1651 (0.0031) & 0.0478 (0.0038) \\
S3 LocalLate & 0.1102 (0.0049) & 0.0000 (0.0000) & -0.0007 (0.0005) & 0.0000 (0.0000) & 0.0000 (0.0000) & 0.0635 (0.0033) & 0.0445 (0.0050) \\
S4 OppositePatches & 0.0843 (0.0098) & -0.0000 (0.0000) & -0.0004 (0.0005) & 0.0000 (0.0000) & 0.0000 (0.0000) & 0.1352 (0.0043) & 0.0577 (0.0050) \\
S5 CrossWindowXOR & 0.4138 (0.0115) & 0.0000 (0.0000) & -0.0022 (0.0010) & 0.0000 (0.0000) & 0.0000 (0.0000) & 0.2953 (0.0117) & 0.0401 (0.0137) \\
S6 Null & -0.0027 (0.0016) & 0.0113 (0.0041) & 0.0064 (0.0046) & 0.0005 (0.0020) & -0.0011 (0.0013) & 0.0089 (0.0036) & 0.0213 (0.0033) \\
S7 MixedLocalGlobal & 0.1220 (0.0081) & 0.0000 (0.0000) & -0.0020 (0.0017) & 0.0000 (0.0000) & 0.0000 (0.0000) & 0.1266 (0.0041) & 0.0853 (0.0057) \\
S8 ImbalancedLocal & 0.1446 (0.0070) & 0.0000 (0.0000) & -0.0006 (0.0010) & 0.0000 (0.0000) & 0.0000 (0.0000) & 0.1265 (0.0055) & 0.1357 (0.0090) \\
S9 WeakTransient & 0.1125 (0.0090) & -0.0008 (0.0005) & -0.0012 (0.0014) & 0.0009 (0.0004) & 0.0016 (0.0011) & 0.1540 (0.0055) & 0.1937 (0.0061) \\
S10 LowVarianceLocal & 0.0967 (0.0098) & 0.0000 (0.0000) & -0.0023 (0.0020) & 0.0000 (0.0000) & 0.0002 (0.0001) & 0.0947 (0.0077) & 0.2847 (0.0081) \\
S11 PhaseShift & 0.1165 (0.0051) & -0.0003 (0.0002) & -0.0033 (0.0011) & 0.0012 (0.0006) & -0.0006 (0.0005) & 0.1664 (0.0046) & 0.1816 (0.0054) \\
S12 DistributedTransients & 0.1624 (0.0065) & -0.0000 (0.0000) & -0.0016 (0.0010) & 0.0000 (0.0000) & 0.0000 (0.0000) & 0.2310 (0.0055) & 0.2248 (0.0061) \\
\bottomrule
\end{tabular}}
\caption{Simulation study: mean paired difference in held-out Brier score (competitor minus GL-FRF) with its Monte Carlo standard error in parentheses. Positive values favour GL-FRF.}
\label{tab:sim_paired}
\end{sidewaystable}

%% file: tables/Table_Data_Audit.tex
\begin{sidewaystable}[p]
\centering
\small
\setlength{\tabcolsep}{4pt}
\resizebox{\linewidth}{!}{\begin{tabular}{lrrrrrrrrrl}
\toprule
Dataset & Train & Test & $p$ & $C$ & Min.\ class & $\widehat\tau_0$ & $\widehat K_0$ & $\widehat W$ & $\bar a_0$ & Splits \\
\midrule
ChlorineConcentration & 467 & 3840 & 166 & 3 & 91 & 0.99 & 11 & 10 & 0.00 & 5 \\
Coffee & 28 & 28 & 286 & 2 & 14 & 0.80 & 3 & 20 & 0.40 & 5 \\
Computers & 250 & 250 & 720 & 2 & 125 & 0.99 & 62 & 20 & 0.12 & 5 \\
ECG200 & 100 & 100 & 96 & 2 & 31 & 0.95 & 12 & 10 & 0.76 & 5 \\
Earthquakes & 322 & 139 & 512 & 2 & 58 & 0.95 & 5 & 10 & 0.18 & 5 \\
Fish & 175 & 175 & 463 & 7 & 21 & 0.95 & 12 & 5 & 0.59 & 5 \\
FordA & 3601 & 1320 & 500 & 2 & 1755 & 0.99 & 73 & 5 & 0.00 & 1 \\
FreezerRegularTrain & 150 & 2850 & 301 & 2 & 75 & 0.99 & 20 & 10 & 0.00 & 5 \\
GunPoint & 50 & 150 & 150 & 2 & 24 & 0.95 & 6 & 20 & 0.00 & 5 \\
Ham & 109 & 105 & 431 & 2 & 52 & 0.80 & 12 & 10 & 0.77 & 5 \\
Herring & 64 & 64 & 512 & 2 & 25 & 0.99 & 31 & 10 & 0.06 & 5 \\
Meat & 60 & 60 & 448 & 3 & 20 & 0.95 & 3 & 10 & 0.19 & 5 \\
PowerCons & 180 & 180 & 144 & 2 & 90 & 0.80 & 9 & 20 & 0.17 & 5 \\
ProximalPhalanxOutlineCorrect & 600 & 291 & 80 & 2 & 194 & 0.95 & 8 & 5 & 0.76 & 5 \\
Strawberry & 613 & 370 & 235 & 2 & 219 & 0.99 & 11 & 20 & 0.41 & 5 \\
ToeSegmentation1 & 40 & 228 & 277 & 2 & 20 & 0.90 & 14 & 20 & 0.30 & 5 \\
Trace & 100 & 100 & 275 & 4 & 21 & 0.99 & 15 & 5 & 0.00 & 5 \\
Wine & 57 & 54 & 234 & 2 & 27 & 0.99 & 5 & 10 & 0.60 & 5 \\
WormsTwoClass & 181 & 77 & 900 & 2 & 76 & 0.95 & 29 & 5 & 0.28 & 5 \\
Yoga & 300 & 3000 & 426 & 2 & 137 & 0.95 & 7 & 5 & 0.41 & 5 \\
\bottomrule
\end{tabular}}
\caption{UCR datasets: official training and test sizes, series length $p$, number of classes $C$ and smallest training class; median selected threshold $\widehat\tau_0$ and number of components $\widehat K_0$ of the global expert, median number of windows $\widehat W$, mean global weight $\bar a_0$ of GL-FRF over splits, and number of analysed splits.}
\label{tab:audit}
\end{sidewaystable}

%% file: tables/Table_UCR_Brier.tex
\begin{sidewaystable}[p]
\centering
\scriptsize
\setlength{\tabcolsep}{2.5pt}
\resizebox{\linewidth}{!}{\begin{tabular}{lrrrrrrrrrrrr}
\toprule
Dataset & GL-FRF & WC-FRF & Stack-0 & WC-0 & Scr-stack & G-tuned & G-tuned-bud & G-99 & L-eq & IF-mean & IF-TSF & FkNN
\\
\midrule
ChlorineConcentration & 0.410 (0.011) & \textbf{0.362 (0.012)} & 0.410 (0.011) & 0.379 (0.011) & 0.459 (0.018) & 0.503 (0.017) & 0.501 (0.018) & 0.501 (0.018) & 0.465 (0.008) & 0.489 (0.007) & 0.455 (0.011) & 0.613 (0.005) \\
Coffee & 0.108 (0.035) & \textbf{0.076 (0.026)} & 0.116 (0.058) & 0.113 (0.022) & 0.161 (0.046) & 0.162 (0.053) & 0.161 (0.048) & 0.226 (0.030) & 0.237 (0.019) & 0.183 (0.021) & 0.138 (0.020) & 0.078 (0.096) \\
Computers & 0.427 (0.026) & 0.440 (0.035) & 0.429 (0.028) & 0.435 (0.027) & 0.473 (0.010) & 0.475 (0.011) & 0.471 (0.009) & 0.477 (0.012) & 0.432 (0.015) & 0.487 (0.029) & \textbf{0.375 (0.027)} & 0.555 (0.033) \\
ECG200 & 0.227 (0.018) & 0.216 (0.034) & 0.234 (0.011) & 0.220 (0.028) & 0.221 (0.033) & 0.220 (0.032) & 0.220 (0.028) & 0.258 (0.020) & 0.255 (0.016) & 0.267 (0.025) & 0.232 (0.023) & \textbf{0.210 (0.062)} \\
Earthquakes & 0.376 (0.017) & 0.370 (0.013) & 0.375 (0.016) & 0.367 (0.012) & 0.391 (0.021) & 0.409 (0.018) & 0.408 (0.022) & 0.398 (0.018) & 0.381 (0.009) & 0.372 (0.007) & \textbf{0.365 (0.009)} & 0.401 (0.015) \\
Fish & 0.387 (0.004) & 0.334 (0.037) & 0.391 (0.005) & 0.356 (0.023) & 0.381 (0.014) & 0.386 (0.015) & 0.380 (0.015) & 0.401 (0.015) & 0.459 (0.028) & 0.509 (0.024) & 0.417 (0.031) & \textbf{0.332 (0.026)} \\
FordA & 0.244 & \textbf{0.239} & 0.244 & 0.244 & 0.268 & 0.268 & 0.266 & 0.271 & 0.244 & 0.399 & 0.395 & 0.354 \\
FreezerRegularTrain & \textbf{0.023 (0.004)} & 0.031 (0.009) & 0.024 (0.004) & 0.049 (0.007) & 0.157 (0.040) & 0.203 (0.056) & 0.202 (0.057) & 0.202 (0.058) & 0.206 (0.045) & 0.182 (0.019) & 0.024 (0.004) & 0.321 (0.011) \\
GunPoint & \textbf{0.083 (0.035)} & 0.101 (0.046) & \textbf{0.083 (0.035)} & 0.111 (0.033) & 0.219 (0.023) & 0.247 (0.039) & 0.241 (0.033) & 0.238 (0.033) & 0.221 (0.012) & 0.201 (0.018) & 0.126 (0.019) & 0.186 (0.043) \\
Ham & 0.326 (0.030) & 0.336 (0.038) & 0.339 (0.039) & 0.338 (0.030) & 0.331 (0.031) & 0.322 (0.026) & \textbf{0.314 (0.021)} & 0.393 (0.030) & 0.374 (0.027) & 0.372 (0.014) & 0.349 (0.042) & 0.457 (0.185) \\
Herring & 0.493 (0.035) & 0.495 (0.051) & 0.500 (0.031) & 0.490 (0.038) & 0.480 (0.027) & 0.497 (0.034) & 0.498 (0.033) & 0.491 (0.019) & 0.481 (0.021) & 0.500 (0.050) & \textbf{0.479 (0.051)} & 0.491 (0.019) \\
Meat & 0.074 (0.033) & 0.061 (0.033) & 0.078 (0.034) & 0.069 (0.027) & 0.090 (0.031) & 0.084 (0.028) & 0.083 (0.029) & 0.118 (0.043) & 0.140 (0.042) & 0.094 (0.023) & 0.070 (0.027) & \textbf{0.027 (0.060)} \\
PowerCons & 0.028 (0.007) & \textbf{0.017 (0.005)} & 0.027 (0.007) & 0.052 (0.004) & 0.086 (0.030) & 0.058 (0.017) & 0.058 (0.017) & 0.107 (0.012) & 0.203 (0.011) & 0.030 (0.011) & 0.027 (0.006) & 0.047 (0.012) \\
ProximalPhalanxOutlineCorrect & 0.245 (0.020) & \textbf{0.226 (0.017)} & 0.248 (0.019) & 0.230 (0.016) & 0.246 (0.019) & 0.247 (0.021) & 0.246 (0.020) & 0.248 (0.010) & 0.289 (0.021) & 0.291 (0.019) & 0.245 (0.022) & 0.245 (0.014) \\
Strawberry & 0.066 (0.008) & \textbf{0.057 (0.007)} & 0.065 (0.008) & \textbf{0.057 (0.007)} & 0.080 (0.012) & 0.082 (0.011) & 0.080 (0.011) & 0.081 (0.010) & 0.140 (0.028) & 0.139 (0.012) & 0.096 (0.011) & 0.101 (0.017) \\
ToeSegmentation1 & 0.470 (0.010) & 0.491 (0.027) & 0.476 (0.010) & 0.478 (0.015) & 0.474 (0.016) & 0.483 (0.025) & 0.477 (0.025) & 0.493 (0.016) & 0.459 (0.011) & 0.442 (0.014) & \textbf{0.428 (0.013)} & 0.568 (0.110) \\
Trace & 0.216 (0.035) & 0.142 (0.033) & 0.216 (0.035) & 0.173 (0.022) & 0.395 (0.025) & 0.432 (0.030) & 0.429 (0.029) & 0.424 (0.024) & 0.321 (0.020) & 0.247 (0.022) & \textbf{0.030 (0.006)} & 0.476 (0.123) \\
Wine & 0.270 (0.075) & 0.278 (0.096) & 0.302 (0.103) & 0.263 (0.090) & \textbf{0.243 (0.049)} & 0.249 (0.042) & 0.258 (0.054) & \textbf{0.243 (0.055)} & 0.326 (0.086) & 0.297 (0.087) & 0.279 (0.075) & 0.289 (0.291) \\
WormsTwoClass & 0.404 (0.029) & 0.403 (0.030) & 0.401 (0.046) & 0.403 (0.026) & 0.419 (0.025) & 0.421 (0.028) & 0.418 (0.028) & 0.423 (0.031) & \textbf{0.393 (0.023)} & 0.422 (0.023) & 0.407 (0.020) & 0.494 (0.023) \\
Yoga & 0.268 (0.018) & 0.259 (0.008) & 0.268 (0.018) & 0.260 (0.009) & 0.278 (0.019) & 0.279 (0.019) & 0.277 (0.023) & 0.292 (0.018) & 0.295 (0.009) & 0.262 (0.011) & \textbf{0.243 (0.010)} & 0.284 (0.025) \\
\bottomrule
\end{tabular}}
\caption{UCR datasets: mean (SD) held-out Brier score over the official split and 4 stratified resamples (FordA: official split only). GL-FRF: global--local functional random forest (late fusion); WC-FRF: window-concatenated functional random forest (early fusion, OOB-tuned); Stack-0: GL-FRF without penalties; WC-0: WC-FRF with default hyperparameters; Scr-stack: the same stack, with the $(\widehat W,\widehat\lambda_G,\widehat\lambda_L)$ of GL-FRF, on random non-contiguous partitions of the grid (locality control); G-tuned: tuned global expert alone (100 trees); G-tuned-bud: the same with $100(\widehat W+1)$ trees; G-99: untuned global forest (99\% FPCA); L-eq: equal soft vote of the local experts; IF-mean, IF-TSF: random-interval forests with interval means, and with mean, standard deviation and slope; FkNN: functional $k$-nearest neighbours. Best mean in bold, including methods tied with it at the reported precision.}
\label{tab:ucr_brier}
\end{sidewaystable}

%% file: tables/Table_UCR_BA.tex
\begin{sidewaystable}[p]
\centering
\scriptsize
\setlength{\tabcolsep}{2.5pt}
\resizebox{\linewidth}{!}{\begin{tabular}{lrrrrrrrrrrrr}
\toprule
Dataset & GL-FRF & WC-FRF & Stack-0 & WC-0 & Scr-stack & G-tuned & G-tuned-bud & G-99 & L-eq & IF-mean & IF-TSF & FkNN
\\
\midrule
ChlorineConcentration & 0.617 (0.010) & \textbf{0.678 (0.018)} & 0.617 (0.010) & 0.654 (0.013) & 0.553 (0.023) & 0.516 (0.017) & 0.515 (0.018) & 0.526 (0.016) & 0.534 (0.014) & 0.521 (0.007) & 0.561 (0.013) & 0.378 (0.020) \\
Coffee & 0.967 (0.024) & 0.972 (0.016) & 0.946 (0.067) & 0.987 (0.018) & 0.901 (0.063) & 0.888 (0.074) & 0.896 (0.076) & 0.875 (0.062) & \textbf{0.993 (0.015)} & 0.926 (0.037) & 0.950 (0.039) & 0.966 (0.049) \\
Computers & 0.671 (0.045) & 0.657 (0.053) & 0.671 (0.043) & 0.658 (0.043) & 0.608 (0.015) & 0.579 (0.017) & 0.573 (0.018) & 0.572 (0.019) & 0.642 (0.019) & 0.581 (0.030) & \textbf{0.723 (0.024)} & 0.518 (0.024) \\
ECG200 & 0.821 (0.024) & 0.816 (0.039) & 0.816 (0.031) & 0.810 (0.028) & \textbf{0.849 (0.044)} & \textbf{0.849 (0.044)} & 0.827 (0.043) & 0.798 (0.022) & 0.748 (0.042) & 0.753 (0.024) & 0.794 (0.025) & 0.842 (0.028) \\
Earthquakes & 0.501 (0.002) & 0.508 (0.021) & 0.501 (0.002) & 0.503 (0.009) & 0.495 (0.011) & 0.498 (0.019) & 0.495 (0.016) & \textbf{0.509 (0.019)} & 0.500 (0.000) & 0.501 (0.008) & 0.503 (0.006) & 0.500 (0.000) \\
Fish & 0.794 (0.026) & 0.813 (0.032) & 0.788 (0.023) & \textbf{0.832 (0.043)} & 0.820 (0.056) & 0.783 (0.030) & 0.793 (0.035) & 0.827 (0.026) & 0.809 (0.042) & 0.659 (0.029) & 0.771 (0.032) & 0.784 (0.018) \\
FordA & 0.894 & 0.859 & 0.895 & 0.863 & 0.851 & 0.851 & 0.857 & 0.848 & \textbf{0.896} & 0.773 & 0.798 & 0.743 \\
FreezerRegularTrain & 0.992 (0.002) & 0.989 (0.004) & 0.992 (0.003) & 0.988 (0.003) & 0.903 (0.039) & 0.862 (0.051) & 0.859 (0.053) & 0.858 (0.048) & 0.845 (0.053) & 0.880 (0.007) & \textbf{0.995 (0.003)} & 0.810 (0.044) \\
GunPoint & 0.949 (0.039) & 0.954 (0.028) & 0.949 (0.039) & 0.955 (0.025) & 0.885 (0.029) & 0.859 (0.058) & 0.863 (0.056) & 0.868 (0.055) & 0.928 (0.020) & 0.867 (0.024) & \textbf{0.961 (0.014)} & 0.897 (0.040) \\
Ham & 0.801 (0.044) & 0.771 (0.073) & 0.801 (0.058) & 0.799 (0.064) & 0.792 (0.050) & 0.798 (0.049) & \textbf{0.809 (0.039)} & 0.736 (0.067) & 0.777 (0.042) & 0.744 (0.032) & 0.754 (0.070) & 0.711 (0.075) \\
Herring & 0.466 (0.057) & 0.488 (0.038) & 0.480 (0.043) & 0.496 (0.056) & 0.518 (0.043) & 0.503 (0.013) & 0.496 (0.030) & 0.509 (0.029) & 0.488 (0.042) & 0.511 (0.062) & \textbf{0.547 (0.071)} & 0.507 (0.061) \\
Meat & 0.983 (0.029) & 0.977 (0.019) & 0.977 (0.032) & 0.983 (0.017) & 0.980 (0.018) & 0.970 (0.022) & 0.967 (0.035) & 0.963 (0.030) & \textbf{0.987 (0.014)} & 0.943 (0.035) & 0.973 (0.025) & \textbf{0.987 (0.030)} \\
PowerCons & 0.987 (0.008) & \textbf{0.992 (0.007)} & 0.986 (0.013) & 0.984 (0.014) & 0.968 (0.021) & 0.973 (0.010) & 0.972 (0.011) & 0.970 (0.014) & 0.927 (0.018) & 0.982 (0.012) & 0.984 (0.009) & 0.967 (0.014) \\
ProximalPhalanxOutlineCorrect & 0.776 (0.024) & \textbf{0.799 (0.020)} & 0.766 (0.021) & 0.789 (0.015) & 0.783 (0.023) & 0.775 (0.021) & 0.775 (0.018) & 0.768 (0.017) & 0.705 (0.017) & 0.707 (0.022) & 0.781 (0.026) & 0.753 (0.010) \\
Strawberry & 0.967 (0.007) & 0.970 (0.012) & \textbf{0.971 (0.007)} & \textbf{0.971 (0.010)} & 0.956 (0.014) & 0.948 (0.015) & 0.951 (0.015) & 0.953 (0.010) & 0.936 (0.014) & 0.899 (0.013) & 0.936 (0.016) & 0.944 (0.010) \\
ToeSegmentation1 & 0.593 (0.034) & 0.568 (0.044) & 0.588 (0.020) & 0.579 (0.014) & 0.579 (0.045) & 0.574 (0.036) & 0.577 (0.026) & 0.559 (0.036) & 0.640 (0.026) & 0.634 (0.026) & \textbf{0.658 (0.049)} & 0.557 (0.033) \\
Trace & 0.870 (0.025) & 0.927 (0.035) & 0.870 (0.027) & 0.919 (0.035) & 0.708 (0.049) & 0.663 (0.038) & 0.688 (0.041) & 0.686 (0.029) & 0.813 (0.034) & 0.786 (0.025) & \textbf{0.997 (0.008)} & 0.780 (0.059) \\
Wine & 0.830 (0.083) & 0.793 (0.082) & 0.811 (0.088) & 0.819 (0.077) & 0.844 (0.095) & 0.848 (0.033) & 0.833 (0.035) & 0.841 (0.034) & 0.781 (0.164) & 0.807 (0.105) & 0.830 (0.090) & \textbf{0.856 (0.145)} \\
WormsTwoClass & 0.689 (0.055) & 0.648 (0.061) & 0.660 (0.085) & 0.648 (0.077) & 0.624 (0.039) & 0.660 (0.067) & 0.658 (0.024) & 0.639 (0.063) & \textbf{0.708 (0.067)} & 0.612 (0.038) & 0.640 (0.049) & 0.554 (0.040) \\
Yoga & 0.832 (0.012) & 0.827 (0.017) & 0.832 (0.012) & 0.831 (0.016) & 0.817 (0.021) & 0.808 (0.022) & 0.809 (0.026) & 0.798 (0.020) & 0.823 (0.008) & 0.824 (0.009) & \textbf{0.844 (0.015)} & 0.800 (0.029) \\
\bottomrule
\end{tabular}}
\caption{UCR datasets: mean (SD) held-out balanced accuracy over the official split and 4 stratified resamples (FordA: official split only). GL-FRF: global--local functional random forest (late fusion); WC-FRF: window-concatenated functional random forest (early fusion, OOB-tuned); Stack-0: GL-FRF without penalties; WC-0: WC-FRF with default hyperparameters; Scr-stack: the same stack, with the $(\widehat W,\widehat\lambda_G,\widehat\lambda_L)$ of GL-FRF, on random non-contiguous partitions of the grid (locality control); G-tuned: tuned global expert alone (100 trees); G-tuned-bud: the same with $100(\widehat W+1)$ trees; G-99: untuned global forest (99\% FPCA); L-eq: equal soft vote of the local experts; IF-mean, IF-TSF: random-interval forests with interval means, and with mean, standard deviation and slope; FkNN: functional $k$-nearest neighbours. Best mean in bold, including methods tied with it at the reported precision.}
\label{tab:ucr_ba}
\end{sidewaystable}

%% file: tables/Table_UCR_Tests.tex
\begin{table}[t]
\centering
\small
\resizebox{\linewidth}{!}{\begin{tabular}{lrrrrrr}
\toprule
Competitor & Datasets & Median diff. & GL-FRF better/worse & $V$ & $p$ & $p_{\mathrm{Holm}}$ \\
\midrule
G-tuned-bud & 20 & 0.0146 & 16/ 4 & 186 & 0.001 & 0.006 \\
Scr-stack & 20 & 0.0153 & 16/ 4 & 181 & 0.003 & 0.009 \\
WC-FRF & 20 & -0.0071 &  7/13 & 65 & 0.143 & 0.286 \\
IF-TSF & 20 & 0.0019 & 12/ 8 & 121 & 0.571 & 0.571 \\
Stack-0 & 20 & 0.0015 & 14/ 6 & 170 & 0.014 & -- \\
WC-0 & 20 & -0.0042 &  7/13 & 84 & 0.452 & -- \\
G-tuned & 20 & 0.0167 & 16/ 4 & 190 & 0.001 & -- \\
G-99 & 20 & 0.0293 & 18/ 2 & 200 & 0.000 & -- \\
L-eq & 20 & 0.0514 & 17/ 3 & 195 & 0.000 & -- \\
IF-mean & 20 & 0.0432 & 17/ 3 & 197 & 0.000 & -- \\
FkNN & 20 & 0.0304 & 14/ 6 & 174 & 0.008 & -- \\
\bottomrule
\end{tabular}}
\caption{UCR datasets: median over datasets of the dataset-level mean difference in Brier score (competitor minus GL-FRF; positive values favour GL-FRF), number of datasets on which GL-FRF was better or worse, and two-sided Wilcoxon signed-rank test. The first 4 rows form the family of main comparisons (Holm adjustment); the remaining rows are descriptive.}
\label{tab:ucr_tests}
\end{table}

%% file: tables/Table_Excess.tex
\begin{table}[t]
\centering
\scriptsize
\setlength{\tabcolsep}{3pt}
\resizebox{\linewidth}{!}{\begin{tabular}{lrrrrrrr}
\toprule
Group (settings) & GL-FRF & WC-FRF & Stack-0 & Scr-stack & G-tuned-bud & IF-TSF & FkNN
\\
\midrule
Simulation (12) & \textbf{0.003} (\textbf{0.013}) & 0.127 (0.416) & \textbf{0.003} (0.016) & 0.004 (0.022) & \textbf{0.003} (0.027) & 0.132 (0.298) & 0.112 (0.287) \\
UCR (20) & 0.031 (0.187) & \textbf{0.021} (\textbf{0.112}) & 0.035 (0.187) & 0.067 (0.366) & 0.073 (0.399) & 0.033 (0.156) & 0.100 (0.446) \\
All (32) & \textbf{0.021} (\textbf{0.187}) & 0.060 (0.416) & 0.023 (\textbf{0.187}) & 0.043 (0.366) & 0.047 (0.399) & 0.070 (0.298) & 0.105 (0.446) \\
\bottomrule
\end{tabular}}
\caption{Excess mean Brier score over the best of the twelve methods in each simulated scenario or dataset: mean (maximum) over the settings of each group. Smallest mean and smallest maximum of each row in bold. Descriptive summary.}
\label{tab:excess}
\end{table}